\documentclass[pdflatex,sn-mathphys-num]{sn-jnl}

\usepackage{subcaption}
\usepackage{graphicx}%
\usepackage{multirow}%
\usepackage{amsmath,amssymb,amsfonts}%
\usepackage{amsthm}%
\usepackage{mathrsfs}%
\usepackage[title]{appendix}%
\usepackage{xcolor}%
\usepackage{textcomp}%
\usepackage{manyfoot}%
\usepackage{booktabs}%
\usepackage{algorithm}%
\usepackage{algorithmicx}%
\usepackage{algpseudocode}%
\usepackage{listings}%

\theoremstyle{thmstyleone}%
\theoremstyle{thmstyletwo}%

\theoremstyle{thmstylethree}%

\begin{document}

\title[Article Title]{QSMOTE-PGM/kPGM: QSMOTE Based PGM and kPGM for Imbalanced Dataset Classification}


\author*[1]{\fnm{Bikash~K.} \sur{Behera}}\email{bikas.riki@gmail.com}

\author*[1]{\fnm{Giuseppe} \sur{Sergioli}}\email{giuseppe.sergioli@gmail.com}

\author*[1,2]{\fnm{Roberto} \sur{Giuntini}}\email{giuntini@unica.it}

\affil[1]{\orgname{Università degli Studi di Cagliari}, \orgaddress{\street{Via Is Mirrions}, \city{Cagliari}, \postcode{09123}, \country{Italy}}}

\affil[2]{\orgdiv{Technische Universität München}, \orgname{Institute for Advanced Study (IAS)}, \orgaddress{\street{Lichtenbergstraße 2a}, \city{Garching b. M\"unchen}, \postcode{85748}, \country{Germany}}}


\abstract{{Quantum-inspired machine learning (QiML) employs mathematical principles from quantum theory, such as Hilbert-space representations and quantum state discrimination, to enhance classical learning algorithms. In this work, we investigate the integration of Quantum Synthetic Minority Oversampling Technique (QSMOTE) variants with two quantum-inspired classifiers: the Pretty Good Measurement (PGM) classifier and the kernelized Pretty Good Measurement (KPGM) classifier. We propose and analyze three QSMOTE variants, namely KNN-based, Fidelity-based, and Margin-based QSMOTE, designed to improve minority-class representation in imbalanced datasets through quantum-inspired similarity and sampling mechanisms. A unified theoretical and empirical comparison of PGM and KPGM is presented under amplitude and stereo encoding strategies with multiple quantum copies. Experimental evaluations on the Telco Customer Churn dataset demonstrate that the proposed quantum-inspired approaches consistently outperform a classical Random Forest baseline, particularly in terms of recall and balanced F1-score. Among all configurations, PGM with stereo encoding and $n_{\mathrm{copies}}=2$ achieves the best performance with an accuracy of $0.8512$ and an F1-score of $0.8234$, while KPGM exhibits competitive and more stable behavior across different QSMOTE variants, reaching accuracies of $0.8511$ under stereo encoding and $0.8483$ under amplitude encoding. The results further show that increasing the number of quantum copies systematically improves classification performance, especially for minority-class detection. This work highlights the effectiveness of combining quantum-inspired oversampling and classification strategies for imbalanced learning, while providing practical insights into the complementary strengths of measurement-based and kernel-based quantum-inspired machine learning frameworks.}}




\maketitle

\section{Introduction} \label{QVP:Sec1}

Quantum-inspired machine learning (QiML) has emerged as a promising interdisciplinary field that blends mathematical structures from quantum mechanics such as Hilbert spaces, density matrices, and probabilistic measurements with classical learning tasks. In particular, the idea of framing classification as a quantum state discrimination (QSD) problem has led to a new family of classifiers that simulate quantum protocols using classical data and resources. These quantum-like classifiers operate by encoding classical data into quantum states and applying quantum-inspired decision rules, thereby offering a novel probabilistic geometry for classification \cite{schuld2015introduction, biamonte2017quantum}. A foundational technique in this paradigm is the \textit{Pretty Good Measurement} (PGM), which originates from quantum hypothesis testing. PGM provides an analytically tractable approximation to the optimal quantum measurement that distinguishes among a finite ensemble of quantum states \cite{hausladen1994pretty}. In a classification setting, each class is associated with a quantum centroid (a density matrix), and the prediction rule selects the class whose measurement operator yields the highest Born-rule probability with the input state \cite{sergioli2019new}. This approach has been adapted for classical data by representing feature vectors as normalized pure states and using the PGM to evaluate similarity through trace inner products \cite{Giuntini2023QuantumInspired, Giuntini2023QuantumClassification}. The conceptual foundation for quantum-inspired classification was laid by Gambs \cite{gambs2008}, who first introduced multiclass classification based on QSD. Related developments connecting quantum-inspired classification and many-valued logical frameworks have also been investigated in the context of quantum computational logics \cite{dallachiara2018manyvalued}. Sergioli et al. extended this framework with centroid-based models \cite{sergioli2018}, and later introduced explicit construction of quantum measurements, including Helstrom measurements for binary classification \cite{sergioli2019}. While the Helstrom measurement offers closed-form optimality for two-class problems \cite{helstrom1969}, its extension to multiclass settings typically requires semidefinite programming \cite{eldar2003}. To address this, approximate strategies like PGM have become the preferred choice for tractable multiclass classification, particularly in quantum-inspired settings. Despite its interpretability and theoretical grounding, PGM faces a key challenge: scalability. Implementing PGM in high-dimensional feature spaces requires expensive matrix operations, including the computation of the inverse square root of the average quantum state, which scales cubically in the Hilbert space dimension. This motivates the need for alternative formulations that preserve the theoretical essence of PGM while improving computational efficiency.

To address this, Cruzeiro et al. \cite{cruzeiro2024quantum} introduced the \textit{kernelized Pretty Good Measurement} (kPGM), a reformulation that expresses the PGM classification rule entirely in terms of dot products between training samples. Instead of operating on density matrices in a high-dimensional space, kPGM works with Gram matrices constructed from inner products and performs all computations within the span of training data. This not only allows implementation on classical hardware but also reduces complexity to depend on the number of samples $N$ rather than the feature dimension $D$. {Although PGM and k-PGM are theoretically equivalent under idealized conditions, their practical implementations involve distinct numerical representations and approximations, which may lead to small quantitative differences in empirical accuracy while preserving identical qualitative behavior and decision trends.} Furthermore, kPGM naturally supports extensions to $m$-copy quantum states by raising inner products to the $m$th power, enabling a smooth transition from linear to highly nonlinear classifiers. As $m \to \infty$, kPGM asymptotically approximates a nearest-neighbor classifier based on maximal inner product overlap. This highlights a deep geometric connection between QSD and classical similarity-based learning. While PGM remains more faithful to the quantum-theoretic origins, kPGM offers a scalable surrogate that preserves key geometric and probabilistic properties while improving computational feasibility. 

{However, beyond the choice of classification model itself, practical deployment of quantum-inspired methods raises an additional and independent challenge related to data characteristics.}
While the theoretical underpinnings of quantum-inspired classification are well established, real-world datasets are often highly imbalanced, where one or more minority classes contain significantly fewer samples than others. This imbalance poses a critical challenge for any classifier, as traditional algorithms tend to be biased toward majority classes, leading to poor generalization for rare but important instances \cite{he2009learning, japkowicz2002class}. The phenomenon is particularly severe in cybersecurity, healthcare, and fraud detection, where minority events (e.g., attacks, disease cases, fraudulent transactions) carry the highest importance. Addressing class imbalance is thus essential for reliable learning and decision-making. Classical oversampling techniques such as the Synthetic Minority Oversampling Technique (SMOTE) \cite{chawla2002smote} have been widely used to mitigate imbalance by generating synthetic samples for the minority class through linear interpolation between neighboring samples. However, these methods are limited by their purely geometric nature they fail to incorporate richer statistical and probabilistic relationships between data points, often producing noisy or redundant samples in high-dimensional feature spaces. To overcome these limitations, we introduce Quantum Synthetic Minority Oversampling Technique (QSMOTE) \cite{mohanty2025quantum} and propose its three novel variants KNN-based, Fidelity-based, and Margin-based oversampling, which integrate quantum-inspired principles such as fidelity weighting, centroid-guided generation, and margin-driven filtering. Fidelity-based QSMOTE enhances intra-class compactness by guiding sample generation toward the minority cluster centroid using a fidelity-inspired scaling factor. Margin-based QSMOTE filters synthetic instances near the decision boundary to retain only confidently classifiable samples, improving boundary clarity. Together, these approaches form a unified quantum-inspired framework for structured and balanced data generation.

When coupled with PGM and kPGM classifiers, QSMOTE variants demonstrate significant advantages over classical baselines. Experimental results reveal that both PGM and kPGM consistently outperform the Random Forest (RF) baseline, which represents a strong classical benchmark. Under stereo encoding with $n=2$ quantum copies, PGM achieved the highest overall accuracy (0.8512) and F1-score (0.8234), while kPGM delivered competitive stability with accuracy 0.8511 and F1-score 0.8225, surpassing RF's best F1-score of 0.7516. Furthermore, the quantum-inspired models exhibited remarkable gains in recall, reaching 0.8594 under amplitude encoding, indicating superior capability in identifying minority instances. These findings highlight that quantum-inspired oversampling and classification together provide a powerful synergy: QSMOTE enhances data balance by exploiting quantum geometric similarity, and PGM/kPGM leverage Hilbert-space probability structures to achieve balanced precision–recall trade-offs. The result is a robust and interpretable framework for imbalanced learning, combining the generalization strength of classical ensemble methods with the probabilistic richness of quantum state representations.

\subsection{Contributions}
The main contributions of this paper are as follows:
\begin{itemize}
\item[1)] We propose three novel QSMOTE variants, KNN-based, Fidelity-based, and Margin-based which extend classical oversampling paradigms with quantum-inspired mechanisms such as fidelity-based weighting, centroid-driven compactness, and margin-aware filtering. These techniques effectively mitigate class imbalance by generating synthetic samples that preserve the geometric and probabilistic structure of minority classes within the Hilbert-space representation.

\item[2)] We present a unified quantum-inspired formulation of the proposed methods, analyzing their computational behavior in terms of local versus global sampling dynamics, centroid attraction, and classifier-guided margin filtering. This formulation provides an interpretable bridge between classical oversampling heuristics and quantum geometric similarity, establishing a mathematically grounded framework for balanced data generation.

\item[3)] We conduct comprehensive empirical evaluations using both classical and quantum-inspired classifiers under multiple encoding strategies (amplitude, and stereo) and simulation settings. The results demonstrate that the proposed QSMOTE variants significantly enhance minority class representation and improve the overall generalization ability of the classifiers across diverse imbalanced datasets.

\item[4)] We show that when integrated with PGM and kPGM classifiers, QSMOTE variants yield state-of-the-art performance, outperforming the best classical baseline RF in terms of accuracy, F1-score, and recall. Specifically, PGM achieves an F1-score of 0.8234 and recall of 0.8594 under amplitude encoding, surpassing RF's best F1-score of 0.7516. This establishes a new quantum-inspired benchmark for imbalanced learning.

\item[5)] We highlight that the synergy between quantum-inspired oversampling (QSMOTE) and probabilistic quantum classifiers (PGM/kPGM) offers a scalable, interpretable, and robust alternative to existing classical approaches, paving the way for next-generation quantum-inspired solutions for imbalanced data classification.

\end{itemize}

\subsection{Organization}

The rest of the paper is structured as follows: {Section~\ref{qml_based} introduces the taxonomy of quantum-based machine learning systems together with the associated quantum hardware architecture and computational perspective.} Section~\ref{background} presents the theoretical background of PGM and kPGM classifiers. Section~\ref{SecIII} describes the proposed QSMOTE variants and their underlying methodologies. Section~\ref{SecIV} reports the experimental setup and empirical results under different classifiers and encoding strategies. Finally, Section~\ref{SecV} concludes the paper with discussions, insights, and future research directions.

\section{Quantum-Based Machine Learning Systems}\label{qml_based}

Quantum machine learning (QML) has emerged as an interdisciplinary research area that combines quantum computing principles with machine learning methodologies to improve computational efficiency, feature representation, and learning capability \cite{schuld2015introduction,biamonte2017quantum,schuld2018supervised}. Depending on the role played by quantum computation within the learning pipeline, quantum-based machine learning systems can generally be categorized into three major paradigms: end-to-end QML, hybrid quantum-classical machine learning, and quantum-inspired machine learning. These paradigms differ in terms of hardware dependence, computational workflow, scalability, and implementation complexity.

\subsection{Taxonomy of Quantum-Based Machine Learning Systems}

End-to-end QML refers to frameworks in which the entire learning pipeline, including data encoding, feature transformation, training, and inference, is performed directly on quantum hardware \cite{biamonte2017quantum,schuld2018supervised}. In such systems, classical data are encoded into quantum states through quantum feature maps or state preparation circuits, and the subsequent computation is carried out entirely using quantum operations executed on a Quantum Processing Unit (QPU). Examples include quantum support vector machines, quantum neural networks, variational quantum eigensolvers for learning tasks, and fully quantum kernel estimation methods \cite{havlicek2019supervised,rebentrost2014quantum}. The main motivation behind end-to-end QML is the possibility of achieving computational speedups and enhanced representation power through quantum superposition, entanglement, and interference. However, these approaches remain constrained by current noisy intermediate-scale quantum (NISQ) hardware limitations, including decoherence, gate noise, limited qubit counts, and measurement overhead \cite{preskill2018quantum}. Hybrid quantum-classical machine learning systems combine quantum computational modules with classical optimization and control procedures \cite{cerezo2021variational}. In these systems, a parameterized quantum circuit is executed on a QPU, while parameter optimization is performed using classical optimizers such as gradient descent, COBYLA, SPSA, or Adam. Hybrid QML approaches are currently among the most practical paradigms for near-term quantum computing because they reduce the quantum hardware requirements while still exploiting quantum feature representations and variational learning capabilities. Examples include variational quantum classifiers, quantum convolutional neural networks, quantum generative adversarial networks, and hybrid quantum reinforcement learning systems \cite{farhi2018classification,schuld2020circuit}. In these architectures, the quantum computer typically acts as a trainable nonlinear feature extractor, while classical processors manage optimization, loss evaluation, and parameter updates. Quantum-inspired machine learning represents a distinct paradigm in which mathematical structures and principles originating from quantum theory are employed within classical computational frameworks without requiring execution on quantum hardware \cite{ciliberto2018quantum}. These methods utilize concepts such as Hilbert spaces, density matrices, quantum state discrimination, fidelity measures, tensor-product feature mappings, and probabilistic measurement operators to design enhanced classical learning algorithms. Unlike end-to-end or hybrid QML approaches, quantum-inspired systems are implemented entirely on classical processors or simulators while preserving several geometric and probabilistic advantages motivated by quantum mechanics. The present work belongs to this category of quantum-inspired machine learning. Specifically, the proposed framework employs quantum-inspired oversampling strategies together with PGM and KPGM classifiers to address imbalanced classification problems through Hilbert-space representations and quantum state discrimination principles.

\subsection{Quantum Hardware Architecture and Computational Components}

QML systems that rely on quantum hardware are generally built upon specialized computational architectures centered around the QPU \cite{nielsen2002quantum,preskill2018quantum}. The QPU acts as the core computational component responsible for manipulating qubits through quantum gates, entanglement operations, and measurement processes. Similar to the role of a classical Central Processing Unit (CPU), the QPU performs the fundamental computational operations required for quantum algorithms. Depending on the underlying hardware technology, qubits may be implemented using superconducting circuits, trapped ions, photonic systems, neutral atoms, or spin-based architectures. A typical quantum computing architecture consists of several interconnected components. First, the QPU contains the physical qubits and gate-control mechanisms responsible for quantum state evolution. Second, classical control electronics are required to generate microwave pulses, control gate timing, calibrate qubit operations, and process measurement outcomes. Third, quantum memory architectures have been proposed to facilitate efficient storage and retrieval of quantum information. Among these, conceptual models such as quantum random access memory (qRAM) and Q1RAM have been investigated as mechanisms for enabling high-speed access to quantum data structures and large-scale quantum datasets \cite{giovannetti2008quantum,arunachalam2015guest}. In principle, such memory architectures may support efficient data loading and manipulation within QML workflows, particularly for tasks involving high-dimensional feature representations, amplitude encoding, and quantum linear algebra routines. The interaction between QPUs and quantum memory systems plays an important role in the computational efficiency of QML algorithms. In end-to-end QML systems, quantum hardware may provide advantages for operations involving high-dimensional vector manipulation, inner-product estimation, matrix inversion, and probabilistic sampling \cite{rebentrost2014quantum,lloyd2014quantum}. Quantum parallelism enables a quantum processor to evolve multiple computational basis states simultaneously, while entanglement and interference can enhance feature correlations and nonlinear representations. In hybrid quantum-classical systems, the QPU is typically used for feature extraction or expectation-value computation, whereas optimization and parameter updates are delegated to classical hardware \cite{cerezo2021variational}. Consequently, one of the major motivations for QML is the possibility of reducing the computational complexity of specific learning tasks \cite{biamonte2017quantum,cerezo2021variational}. Quantum algorithms for kernel estimation, optimization, and linear algebra have demonstrated theoretical speedups under certain assumptions regarding scalable hardware and efficient quantum memory access \cite{havlicek2019supervised,lloyd2014quantum}. Despite these theoretical advantages, practical QML remains constrained by current NISQ hardware limitations, including noise accumulation, restricted qubit connectivity, limited coherence times, and expensive data-loading procedures \cite{preskill2018quantum}. Consequently, many contemporary studies employ quantum-inspired formulations and quantum simulations rather than direct execution on large-scale fault-tolerant quantum hardware. Although end-to-end and hybrid QML systems may rely on specialized quantum hardware components such as QPUs and qRAM/Q1RAM architectures, the present work focuses on a quantum-inspired machine learning framework implemented using classical computational resources and Qiskit-based simulation tools. Therefore, the discussion of quantum hardware in this section is intended to provide conceptual context for quantum-based machine learning systems rather than describe the computational infrastructure directly employed in this study. The proposed QSMOTE-PGM/kPGM framework does not claim hardware-level quantum advantage, but instead investigates how quantum-theoretic principles can enhance imbalanced data classification within practically realizable computational environments.\color{black}

\section{Brief Introduction to PGM and kPGM}\label{background}

Quantum-inspired classifiers based on QSD have emerged as powerful alternatives to conventional machine learning (ML) models. They offer a fundamentally different view of classification through the probabilistic and geometric structure of quantum measurement theory. Among these, the PGM has gained particular prominence for its elegant mathematical form and near-optimal discrimination capability between nonorthogonal quantum states. Recent developments have introduced the kPGM, which reformulates PGM entirely in terms of inner products between samples. This enables scalable and efficient implementation on classical hardware while preserving the Hilbert-space geometry that underlies QSD. The kPGM thus bridges the gap between quantum-theoretic classification and kernel-based learning, making it a suitable backbone for hybrid schemes such as QSMOTE-assisted quantum-inspired classifiers. In the context of imbalanced datasets, PGM and kPGM are particularly attractive because they inherently measure probabilistic overlaps between states (via the Born rule), allowing minority-class samples to influence the classification boundary even when numerically underrepresented.

\subsection{Pretty Good Measurement (PGM) Classifier}

The PGM classifier is a quantum-inspired algorithm for multiclass classification, grounded in the principles of QSD. It classifies classical data by encoding each feature vector into a quantum state and distinguishing these states using the PGM strategy. Each classical feature vector $\vec{x} \in \mathbb{R}^d$ is mapped to a quantum state represented by a density matrix:
\begin{eqnarray}
\vec{x} \mapsto \rho_{\vec{x}}, \quad \text{where } \rho_{\vec{x}} \succeq 0, \quad \operatorname{Tr}(\rho_{\vec{x}}) = 1.
\end{eqnarray}
Common encoding methods include amplitude encoding, and stereographic projection, which preserve geometric similarity within the Hilbert space. For a dataset with classes $i \in {1, 2, \ldots, \ell}$, the quantum centroid (average density matrix) for class $i$ is defined as:
\begin{eqnarray}
\rho^{(i)} = \frac{1}{|S^{(i)}|} \sum_{\vec{x_j} \in S^{(i)}} \rho_{\vec{x_j}},
\end{eqnarray}
where $S^{(i)}$ denotes the set of samples from class $i$.
To enhance class separability, the data can be “lifted” into higher-dimensional spaces using $n$ tensor copies (analogous to nonlinear feature mapping):
\begin{eqnarray}
\rho^{(i)}_{(n)} = \frac{1}{|S^{(i)}|} \sum_{\vec{x_j} \in S^{(i)}} \rho_{\vec{x_j}}^{\otimes n}.
\end{eqnarray}

The ensemble of class states is represented as:

\begin{eqnarray}
\mathcal{R} = \left\{ (p_i, \rho^{(i)}_{(n)}) \right\}_{i=1}^\ell,
\end{eqnarray}
where $p_i = \frac{|S^{(i)}|}{\sum_j |S^{(j)}|}$ is the empirical prior probability of class $i$.
The average quantum state is then given by:
\begin{eqnarray}
\sigma = \sum_{i=1}^\ell p_i \rho^{(i)}_{(n)}.
\end{eqnarray}

The measurement operators (Positive Operator-Valued Measure (POVM) elements) used in PGM are:
\begin{eqnarray}
E_i = \sigma^{-1/2}  p_i \rho^{(i)}_{(n)}  \sigma^{-1/2}, \label{eq6}
\end{eqnarray}
where $\sigma^{-1/2}$ denotes the pseudoinverse of $\sigma$.
If $\sigma$ is not full-rank, the adjusted operators
\begin{eqnarray}
F_i = E_i + \frac{1}{\ell} P_{\ker(\sigma)},
\end{eqnarray}
are used, where $P_{\ker(\sigma)}$ projects onto the kernel of $\sigma$ to ensure $\sum_i F_i = \mathbb{I}$. For a new test sample $\vec{x}$, encoded as $\rho_{\vec{x}}^{\otimes n}$, the score function for class $i$ is:
\begin{eqnarray}
f(\vec{x})_i = \operatorname{Tr}(F_i \rho_{\vec{x}}^{\otimes n}),
\end{eqnarray}
and the predicted label is:
\begin{eqnarray}
\text{label}(\vec{x}) = \arg\max_{i} f(\vec{x})_i.
\end{eqnarray}

The success probability of PGM satisfies:
\begin{eqnarray}
P_{\text{PGM}}^2 \leq P_{\text{OPT}} \leq P_{\text{PGM}},
\end{eqnarray}
where $P_{\text{OPT}}$ denotes the optimal discrimination probability. The analytical bound for evaluation is:
\begin{eqnarray}
\text{PGM}_b(\mathcal{R}, n) = \sum_{i=1}^\ell p_i \operatorname{Tr}(F_i \rho^{(i)}_{(n)}).
\end{eqnarray}

The use of multiple copies ($n > 1$) embeds data into a higher-dimensional Hilbert space, analogous to kernel lifting in classical ML. This typically improves class separation especially for minority classes though at higher computational cost. Empirically, increasing $n$ enhances recall and F1-scores for imbalanced datasets, indicating that the Hilbert-space feature lifting of PGM is especially effective when combined with QSMOTE-generated balanced samples.

\subsection{Kernelized Pretty Good Measurement (kPGM) Classifier}

The kPGM classifier retains the probabilistic foundation of PGM but employs kernel-based reformulation to achieve computational efficiency. Instead of directly manipulating high-dimensional density matrices, all operations are expressed through Gram matrices allowing the model to scale with the number of samples rather than the Hilbert space dimension. Each data vector $x_i$ is first normalized and encoded as a pure quantum state:
\begin{eqnarray}
\rho_i = x_i x_i^\top,
\end{eqnarray}
and the average state is:
\begin{eqnarray}
\rho=\frac{1}{N}\sum_{i=1}^{N} x_i x_i^\top,
\end{eqnarray}
where $x_i \in \mathbb{R}^q$ and $N$ is the number of training samples. The Gram matrix is then defined as:
\begin{eqnarray}
G_{ij} = x_i^\top x_j,
\end{eqnarray}
and the projector onto the data subspace is:
\begin{eqnarray}
\Pi_S = \sum_{i,j} x_i (G^{-1})_{ij} x_j^\top.
\end{eqnarray}

For a query vector $z \in \mathbb{R}^q$, the POVM element for class $k$ is defined as:
\begin{eqnarray}
E_k = \frac{1}{N} \sum_{i \in C_k} \rho^{-1/2} x_i x_i^\top \rho^{-1/2},
\end{eqnarray}
where $C_k$ denotes the subset of training samples belonging to class $k$.
The conditional probability of observing class $k$ given $z$ is:
\begin{align}
\Pr(k|z) &= z^\top E_k z \nonumber \\
&= \frac{1}{N} \sum_{i \in C_k} z^\top \rho^{-1/2} x_i x_i^\top \rho^{-1/2} z.
\end{align}

By expressing these operators entirely in terms of Gram matrices, the classification rule simplifies to:
\begin{equation}
\Pr(k|z) = w^\top G^{-1/2} \Pi_k G^{-1/2} w,
\end{equation}
where $w_i = x_i^\top z$ and $\Pi_k$ is the projection matrix selecting samples of class $k$.
This compact kernel-based representation ensures that all computations depend only on pairwise inner products, making kPGM scalable and implementable with standard kernel methods. To enhance nonlinearity, kPGM can operate on $m$-fold tensor products:
\begin{eqnarray}
x_i^{(m)} = x_i^{\otimes m}, \quad 
G_{ij}^{(m)} = (x_i^\top x_j)^m, \quad 
w_i^{(m)} = (x_i^\top z)^m,
\end{eqnarray}
allowing quantum-inspired higher-order mappings without explicitly constructing the exponentially large feature space.
As $m \to \infty$, kPGM smoothly converges to a 1-nearest-neighbor classifier:
\begin{eqnarray}
\Pr(k|z) \to \max_{i \in C_k} |x_i^\top z|^{2m},
\end{eqnarray}
and the predicted label is assigned as:
\begin{eqnarray}
\hat{y} = \arg\max_k \Pr(k|z).
\end{eqnarray}

The kPGM classifier thus unifies the probabilistic optimality of quantum measurements with the computational efficiency of kernel methods, making it highly effective for QSMOTE-balanced datasets. Its empirical robustness on benchmark datasets demonstrates how quantum-inspired geometry can enhance minority-class sensitivity and outperform traditional baselines like RF in both accuracy and recall.

\section{Methodology}\label{SecIII}
{In this work, QSMOTE is employed as a data-level preprocessing strategy that operates prior to classifier training. Specifically, minority-class samples are augmented in the original feature space using QSMOTE or its proposed variants, yielding a balanced training dataset. The resulting dataset is then encoded into quantum-inspired representations and used to train either the PGM or k-PGM classifier. In this way, QSMOTE and quantum-inspired classifiers play complementary roles: QSMOTE addresses data imbalance, while PGM and k-PGM perform classification on the balanced dataset using Hilbert-space–based similarity and probabilistic decision rules.}
Here, we propose three novel variants of QSMOTE aimed at improving data balance in imbalanced classification problems. These methods are: 
\begin{itemize}
\item[A.] {KNN-Based QSMOTE}, 
\item[B.] {Fidelity-Based QSMOTE},
\item[C.] {Margin-Based QSMOTE}. 
\end{itemize}
Each variant is motivated by quantum computing concepts, such as fidelity (a measure of quantum state similarity) and angle-based rotation, but is implemented in a classical environment.

\subsection{KNN-Based QSMOTE}

The KNN-Based QSMOTE variant is inspired by classical SMOTE, where synthetic samples are generated by linear interpolation between minority class instances and their nearest neighbors. The process is as follows:

\begin{enumerate}
    \item Identify all samples belonging to the minority class.
    \item Fit a $k$-Nearest Neighbors model using only the minority samples.
    \item For each synthetic instance to be generated:
    \begin{itemize}
        \item Randomly select a minority sample $x_i$.
        \item Find its $k$ nearest neighbors and randomly select one neighbor $x_j$.
        \item Generate the synthetic point using:
        \begin{eqnarray}
        \tilde{x} = x_i + \lambda (x_j - x_i)
        \end{eqnarray}
        where $\lambda$ is a scaling factor derived from an angle or chosen randomly.
    \end{itemize}
\end{enumerate}

This method is simple and preserves local linearity but may suffer from over-generalization in sparse minority regions.

\subsection{Fidelity-Based QSMOTE}

This variant guides the generation of synthetic samples toward the cluster centroid of the minority class, scaled by a fidelity-inspired similarity measure.

\begin{enumerate}
    \item Apply $k$-Means clustering on the dataset.
    \item For each minority class sample $x$ within a cluster, compute the cluster centroid $c$.
    \item Compute the direction vector $\vec{d} = c - x$.
    \item Compute the fidelity-inspired similarity:
    \begin{eqnarray}
    F(x, c) = \left( \frac{x \cdot c}{\|x\| \|c\|} \right)^2
    \end{eqnarray}
    \item Generate the synthetic sample as:
    \begin{eqnarray}
    \tilde{x} = x + \lambda \cdot F(x, c) \cdot \frac{\vec{d}}{\|\vec{d}\|}
    \end{eqnarray}
    where $\lambda$ is a step factor.
\end{enumerate}

This method emphasizes structured generation toward denser regions, promoting compactness and better integration with the existing minority distribution.

\subsection{Margin-Based QSMOTE}

Margin-Based QSMOTE enhances a base QSMOTE variant (e.g., KNN or Fidelity) by filtering synthetic samples that fall near the decision boundary, reducing class ambiguity.

\begin{enumerate}
    \item Use a base QSMOTE variant to generate synthetic samples.
    \item Train a probabilistic classifier (e.g., logistic regression) on the original dataset.
    \item For each synthetic point, compute the class probability $P(y=1 \mid x)$.
    \item Retain only samples satisfying:
    \begin{eqnarray}
    \left|P(y=1 \mid x) - 0.5\right| > \text{margin}
    \end{eqnarray}
\end{enumerate}

This filtering ensures that only confidently classified synthetic points are included, thus improving decision boundary clarity.

\begin{figure*}[htbp]
    \centering
    \begin{subfigure}{0.32\textwidth}
        \centering
        \includegraphics[width=\linewidth]{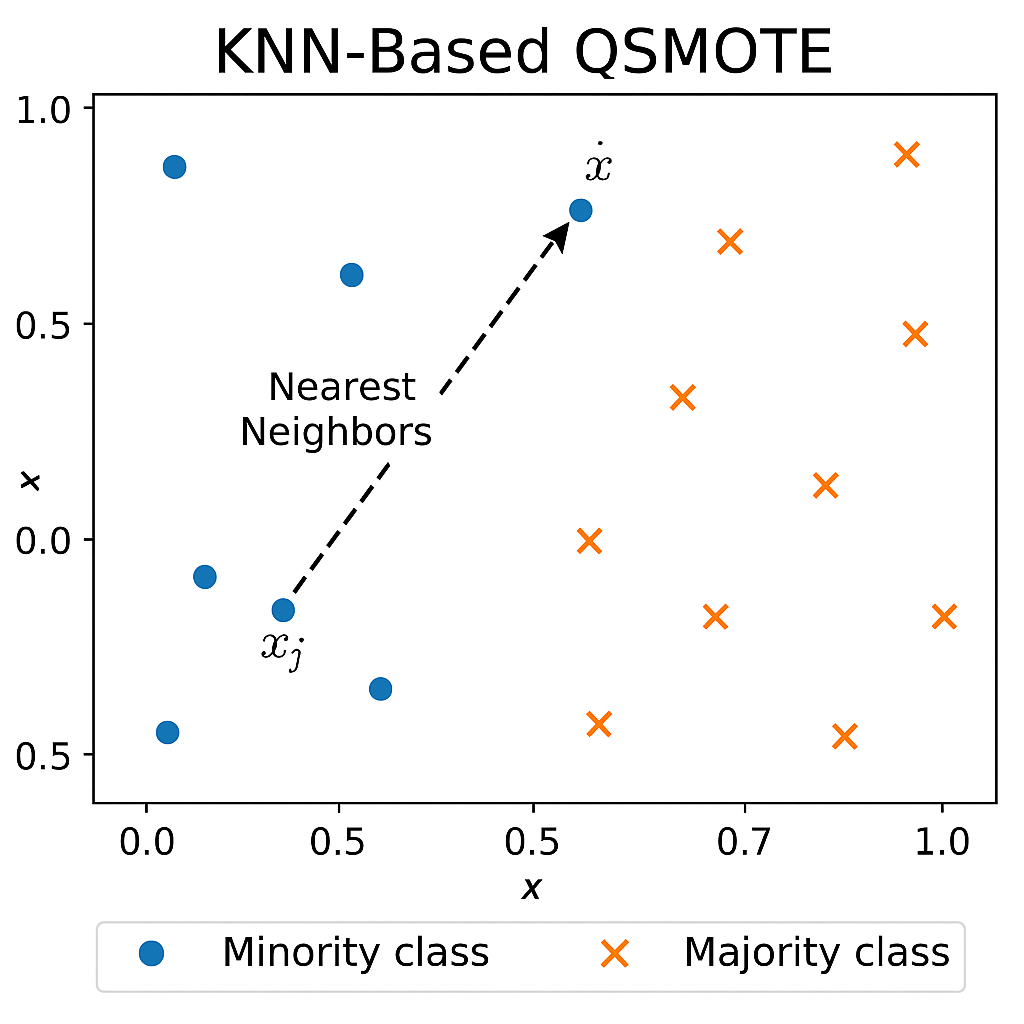}
        \caption{}
        \label{fig:knn_qsmote}
    \end{subfigure}
    \hfill
    \begin{subfigure}{0.32\textwidth}
        \centering
        \includegraphics[width=\linewidth]{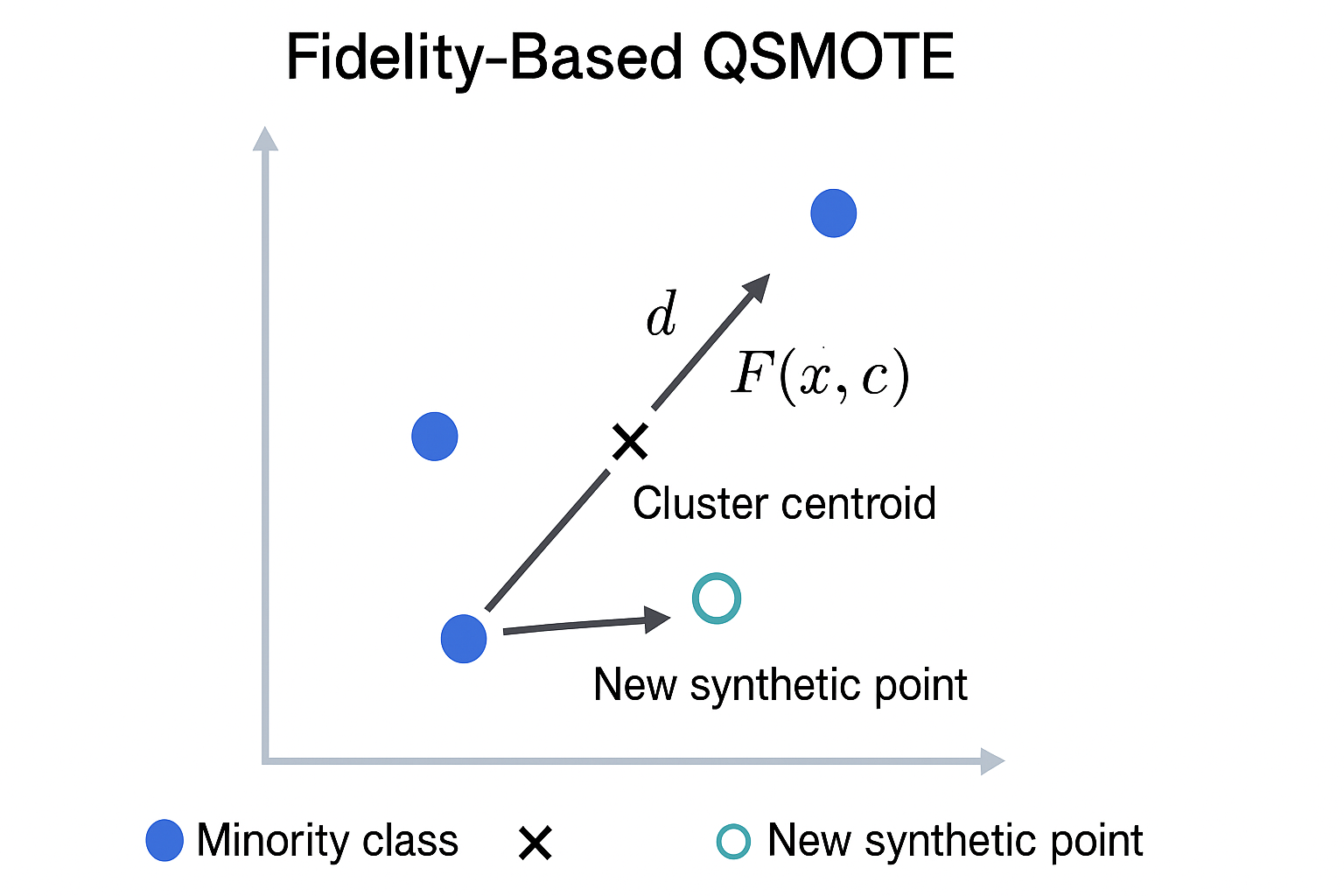}
        \caption{}
        \label{fig:fidelity_qsmote}
    \end{subfigure}
    \hfill
    \begin{subfigure}{0.32\textwidth}
        \centering
        \includegraphics[width=\linewidth]{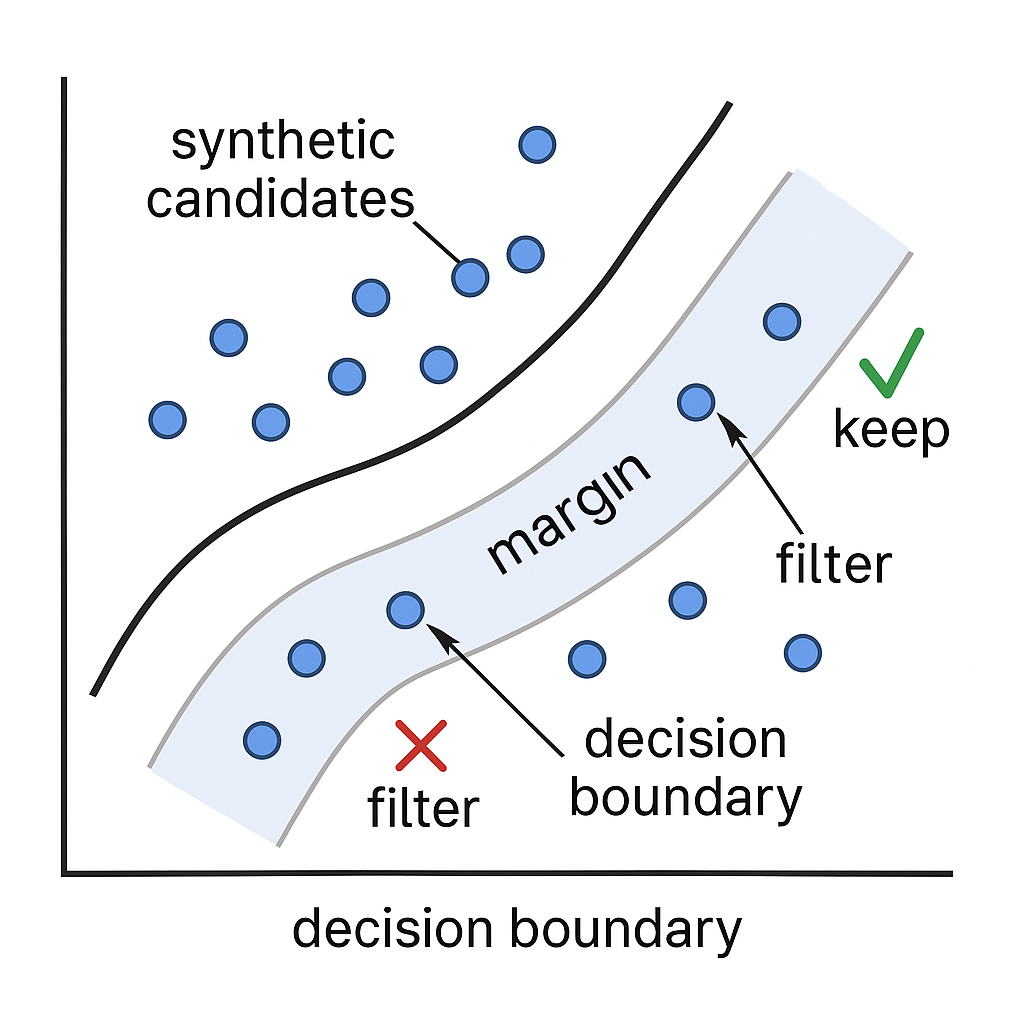}
        \caption{}
        \label{fig:margin_qsmote}
    \end{subfigure}
    \caption{Schematic illustration of the three proposed QSMOTE variants. (a) KNN-based QSMOTE interpolates between a sample and its nearest neighbor. (b) Fidelity-based QSMOTE generates samples directed toward the cluster centroid using fidelity weighting. (c) Margin-based QSMOTE filters synthetic points near the decision boundary to retain only confident samples.}
    \label{fig:qsmote_variants}
\end{figure*}

\begin{table*}[h!]
\centering
\scriptsize
\begin{tabular}{|p{3cm}|p{4cm}|p{4cm}|p{4cm}|}
\hline
\textbf{Feature} & \textbf{KNN-Based QSMOTE} & \textbf{Fidelity-Based QSMOTE} & \textbf{Margin-Based QSMOTE} \\
\hline
Guiding Principle / Goal & Interpolate between a minority point and its nearest neighbors & Generate samples toward cluster centroid, weighted by fidelity similarity & Retain only confidently classifiable synthetic samples \\
\hline
Neighbor / Cluster / Classifier Involvement & Uses $k$-nearest neighbors of a sample, no classifier needed & Requires clustering (e.g., $k$-means) to find centroids, no classifier needed & Requires a classifier (e.g., logistic regression) to estimate confidence \\
\hline
Synthetic Sample Generation & Direct interpolation between neighbors & Generated along centroid direction, scaled by fidelity & Filters synthetic samples after generation based on decision boundary margin \\
\hline
Interpolation / Generation Direction & Between two samples: $x_i$ and $x_j$ & From sample $x$ to cluster centroid $c$ & Not defined; filtering occurs post-generation \\
\hline
Weight / Scaling Factor & $\lambda$ is angle or scaling factor & $\lambda$ is angle/scaling factor, fidelity-weighted: $F(x,c)$ & Not applicable (filtering-based method) \\
\hline
Filtering Criterion & All generated points are kept & None (all generated samples retained) & Samples close to decision boundary (low confidence) are discarded \\
\hline
Effect on Data Distribution & Samples spread along pairwise lines, preserving local geometry & Samples pulled inward toward centroid, encouraging cluster compactness & Reduces ambiguity and noise near decision boundary, improves clarity \\
\hline
Focus & Data geometry and local interpolation & Fidelity-guided directional sampling and global cluster structure & Classifier-driven confidence and margin-based filtering \\
\hline
Global vs Local Sampling Strategy & Local sampling strategy & Global cluster-aware strategy & Filtering based on classifier confidence \\
\hline
\end{tabular}
\caption{Comparison of KNN-Based, Fidelity-Based, and Margin-Based QSMOTE variants}
\label{tab:qsmote_comparison}
\end{table*}

Fig.~\ref{fig:qsmote_variants} provides a schematic comparison of the three proposed QSMOTE variants. In the KNN-based method (Fig.~\ref{fig:knn_qsmote}), synthetic samples are generated along the line segment connecting a minority sample to one of its nearest neighbors, thereby preserving local geometric structure but risking noisy generation in sparse regions. The Fidelity-based method (Fig.~\ref{fig:fidelity_qsmote}) instead directs generation toward the cluster centroid, with the step size modulated by a fidelity-inspired weight, which encourages compact clusters and reduces dispersion. Finally, the Margin-based method (Fig.~\ref{fig:margin_qsmote}) incorporates classifier feedback by filtering synthetic candidates near the decision boundary, retaining only confidently classifiable samples and thereby reducing overlap between classes. Together, these variants highlight three complementary strategies-local interpolation, centroid-guided compactness, and margin-based filtering-for addressing class imbalance in a quantum-inspired framework.
Table~\ref{tab:qsmote_comparison} summarizes the conceptual differences among KNN-Based, Fidelity-Based, and Margin-Based QSMOTE variants. The KNN approach follows a local sampling strategy, where synthetic points are generated by interpolating between a minority sample and its nearest neighbors. This preserves local geometry but risks generating noisy points if neighbors lie near class boundaries. By contrast, the Fidelity-based method leverages clustering and generates synthetic points directed toward the cluster centroid, scaled by a fidelity-weighted factor. This introduces a global, structure-aware perspective that encourages compactness and reduces dispersion of minority points. The Margin-based variant adopts a classifier-driven perspective: synthetic samples are generated first, but only those lying away from the decision boundary (i.e., high-confidence samples) are retained. This reduces overlap and ambiguity near margins, thereby enhancing clarity in the decision space. Overall, KNN-based QSMOTE emphasizes local interpolation, Fidelity-based QSMOTE integrates global cluster information, and Margin-based QSMOTE applies classifier feedback to filter samples, resulting in three complementary strategies for addressing class imbalance.

\begin{algorithm}
\caption{Quantum-Inspired Classification Using PGM or Helstrom Observables}
\label{alg:pgmhqc}
\begin{algorithmic}[1]
\Require Training data $X$, labels $y$, parameters: \textit{rescale}, \textit{encoding}, \textit{n\_copies}, \textit{measure}, \textit{class\_weight}, \textit{n\_splits}, \textit{dtype}
\Ensure Trained quantum-inspired classifier model (centroids + measurement operators)

\State \textbf{Initialize:} set encoding $\in\{\text{amplitude},\text{stereographic}\}$, $n\_copies$, rescaling factor, measurement type $\in\{\text{pgm},\text{hels}\}$, class weights, and numeric precision (\textit{dtype})

\ForAll{class labels $k \in \mathcal{Y}$}
    \State $X_k \Leftarrow \{x \in X \mid y(x)=k\}$
    \ForAll{$x \in X_k$}
        \State \textbf{Encode:} $\rho_x \Leftarrow \mathrm{Encode}(x;\ \textit{encoding})$
        \If{\textit{rescale} is enabled}
            \State \textbf{Rescale:} $\rho_x \Leftarrow \mathrm{Rescale}(\rho_x;\ \textit{rescale})$
        \EndIf
        \State \textbf{Tensorize:} $\tilde{\rho}_x \Leftarrow \rho_x^{\otimes n\_copies}$
    \EndFor
    \State \textbf{Average:} $\rho_k \Leftarrow \frac{1}{|X_k|}\sum_{x\in X_k} \tilde{\rho}_x$
\EndFor

\If{\textit{measure} = \texttt{"pgm"}}
    \State Construct PGM measurement operators $\{\Pi_k\}$ from $\{\rho_k\}$ and \textit{class\_weight}
\ElsIf{\textit{measure} = \texttt{"hels"}}
    \State Construct Helstrom observable $H$ (binary classification only) from $\{\rho_k\}$ and \textit{class\_weight}
\EndIf

\State \textbf{Store:} save $\{\rho_k\}$ and measurement operators (PGM $\{\Pi_k\}$ or Helstrom $H$) for prediction
\State \Return trained model
\end{algorithmic}
\end{algorithm}

\begin{algorithm}
\caption{Kernelized Pretty Good Measurement Classification (KPGM)}
\label{alg:kpgm}
\begin{algorithmic}[1]
\Require Training data $X$, labels $y$, parameters: \textit{n\_copies}, \textit{encoding}, \textit{rescale}, \textit{dtype}, \textit{device}
\Ensure Trained KPGM model (encoded data + labels + Gram matrix)

\State \textbf{Initialize:} set $n\_copies$, encoding $\in\{\text{amplitude},\text{stereographic}\}$, rescaling factor, numeric precision (\textit{dtype}), and compute device (\textit{CPU/GPU})

\State \textbf{Preprocess:} convert $X,y$ to torch tensors with type \textit{dtype} and move to \textit{device}

\ForAll{samples $x_i \in X$}
    \If{\textit{encoding} = \texttt{"amplitude"}}
        \State \textbf{Amplitude-encode:} $x'_i \Leftarrow \mathrm{Normalize}([x_i;\ 1])$ \Comment{append bias term, then normalize}
    \ElsIf{\textit{encoding} = \texttt{"stereographic"}}
        \State \textbf{Stereographic-encode:} $x'_i \Leftarrow \mathrm{InvStereo}(x_i)$
    \EndIf
    \If{\textit{rescale} is enabled}
        \State $x'_i \Leftarrow \mathrm{Rescale}(x'_i;\ \textit{rescale})$
    \EndIf
\EndFor

\State \textbf{Setup:} identify number of classes $C \Leftarrow |\mathrm{unique}(y)|$ and form class-wise index sets $\{\mathcal{I}_k\}_{k=1}^C$

\State \textbf{Compute Gram matrix:}
\For{$i = 1$ to $N$}
    \For{$j = 1$ to $N$}
        \State $K_{ij} \Leftarrow \langle x'_i \mid x'_j \rangle$ \Comment{inner product in encoded space}
    \EndFor
\EndFor

\State \textbf{Store:} save $\{x'_i\}_{i=1}^N$, $y$, and $K$ for prediction
\State \Return trained KPGM model
\end{algorithmic}
\end{algorithm}

\section{Experimental Results}\label{SecIV}

\subsection{Experimental Setup}
All experiments are conducted using Python-based implementations, where classical baseline algorithms are evaluated with scikit-learn, and quantum-inspired oversampling methods are implemented with Qiskit. For the quantum components of QSMOTE and its variants, we employ the {AerSimulator} backend provided by Qiskit Aer. The simulator is configured with {method=`automatic'}, which allows Qiskit to select the most efficient simulation technique (statevector, density matrix, stabilizer, or matrix product state) depending on the circuit structure. Each quantum circuit is executed with $1024$ shots to approximate the probability distribution of measurement outcomes, ensuring statistical stability. The results are collected via the {backend.run(qc, shots=1024).result()} function call, and measurement counts are extracted using {result.get\_counts(0)}. This setup guarantees reproducibility of quantum sampling while balancing execution efficiency. The experimental design thus enables a direct comparison between classical oversampling techniques and their quantum-enhanced counterparts under consistent evaluation protocols.

\subsection{Dataset}

For the purpose of evaluating the proposed method, we utilize the {Telco Customer Churn} dataset, which is publicly available on Kaggle \cite{telcochurn}. This dataset is widely used for benchmarking binary classification algorithms in the context of customer retention analysis. It consists of 7,043 customer records with 21 features that include both numerical and categorical variables such as tenure, contract type, payment method, and internet service. The binary target variable is {Churn}, which indicates whether a customer has discontinued the service. For modeling, the label is encoded as 1 for {Yes} (positive class) and 0 for {No} (negative class). This dataset presents a moderate class imbalance, making it a suitable candidate for testing oversampling methods like QSMOTE and its quantum-enhanced variants.

\subsection{Preprocessing}

Before model training, we applied a structured preprocessing pipeline to the Telco Customer Churn dataset. First, the unique customer identifier column ({customerID}) was removed as it held no predictive value. Next, the {TotalCharges} column was converted to a numeric type, and any resulting missing values (NaNs) were dropped. We then performed one-hot encoding on all categorical features to make them suitable for ML algorithms. To ensure numerical stability and consistency across features, we normalized all feature values to the range [0, 1] using {MinMaxScaler}. Finally, for rapid prototyping and efficient experimentation with quantum-inspired models, we limited the dataset to the first 1000 samples.

\subsection{Hyperparameters}
For the QSMOTE variants, we employed distinct configurations tailored to their respective designs. The KNN-based QSMOTE utilized 5 nearest neighbors to generate synthetic minority samples. The Fidelity-based QSMOTE variant applied KMeans clustering with 3 clusters to guide the generation process based on intra-cluster quantum fidelity. The Margin-based QSMOTE was built atop standard QSMOTE, incorporating a margin threshold of 0.1 to guide resampling decisions, and used Logistic Regression as the base classifier for margin estimation. Regarding the classifiers, the RF is used with default settings from {sklearn}, ensuring reproducibility by setting the {random\_state}. The PGM classifier was a custom quantum-inspired probabilistic graphical model that compared class density matrices. Prior to PGM evaluation, PCA with 16 components was applied for dimensionality reduction. To ensure robust evaluation, all experiments were repeated across 10 random seeds.

\subsection{Evaluation Metrics}

To evaluate the performance of the binary classification models used in our experiments, we employed four standard classification metrics: Accuracy, Precision, Recall, and F1 Score. These metrics are computed based on the number of True Positives (TP), True Negatives (TN), False Positives (FP), and False Negatives (FN) identified by the classifier. {Accuracy} measures the proportion of correctly predicted instances out of the total number of instances. It is defined as:
\begin{eqnarray}
\text{Accuracy} = \frac{TP + TN}{TP + TN + FP + FN}
\end{eqnarray}
This metric is useful when the dataset is balanced, as it gives a general sense of prediction correctness. {Precision} is the fraction of relevant instances among the retrieved instances. It quantifies how many of the predicted positive labels are actually correct:
\begin{eqnarray}
\text{Precision} = \frac{TP}{TP + FP}
\end{eqnarray}
A high precision value indicates a low number of false positives, which is critical in domains where false alarms are costly. {Recall}, also known as sensitivity or true positive rate, measures the proportion of actual positives that were correctly identified:
\begin{eqnarray}
\text{Recall} = \frac{TP}{TP + FN}
\end{eqnarray}
High recall is essential in applications where missing positive instances has serious consequences, such as in medical diagnosis. {F1 Score} is the harmonic mean of Precision and Recall. It provides a balanced measure that considers both false positives and false negatives:
\begin{eqnarray}
\text{F1 Score} = 2 \times \frac{\text{Precision} \times \text{Recall}}{\text{Precision} + \text{Recall}}
\end{eqnarray}
This metric is especially useful in imbalanced classification scenarios where neither precision nor recall alone is sufficient to measure performance comprehensively. Collectively, these metrics give a nuanced and comprehensive view of the model's classification performance across different dimensions. The reported values are represented using mean $\pm$ standard deviation across 5-fold cross-validation.

\subsection{Results}

\begin{table*}[htbp]
\centering
\caption{Performance of QSMOTE and its variants with RF classifier. }
\begin{tabular}{lcccc}
\hline
\textbf{QSMOTE Variant} & \textbf{Accuracy} & \textbf{Precision} & \textbf{Recall} & \textbf{F1 Score} \\
\hline
QSMOTE           & $0.7995 \pm 0.0187$ & $0.7726 \pm 0.0369$ & $0.7218 \pm 0.0315$ & $0.7457 \pm 0.0268$ \\
KNN-Based        & $0.7749 \pm 0.0213$ & $0.6442 \pm 0.0308$ & $0.5224 \pm 0.0584$ & $0.5743 \pm 0.0324$ \\
Fidelity-Based   & $0.8191 \pm 0.0260$ & $0.7647 \pm 0.0412$ & $0.7940 \pm 0.0300$ & $0.7785 \pm 0.0298$ \\
Margin-Filtered  & $0.8371 \pm 0.0226$ & $0.8073 \pm 0.0412$ & $0.7043 \pm 0.0398$ & $0.7516 \pm 0.0327$ \\
\hline
\end{tabular}
\label{tab:qsmote_rf}
\end{table*}

\begin{table*}[htbp]
\centering
\caption{Performance of QSMOTE and its variants with the PGM classifier using amplitude encoding for $n\_copies=1,2$.}
\begin{tabular}{lcccc}
\hline
\multicolumn{5}{c}{$\mathbf{n\_copies=1}$} \\
\hline
\textbf{QSMOTE Variant} & \textbf{Accuracy} & \textbf{Precision} & \textbf{Recall} & \textbf{F1 Score} \\
\hline
QSMOTE           & $0.7808 \pm 0.0275$ & $0.8218 \pm 0.0223$ & $0.6384 \pm 0.0595$ & $0.7166 \pm 0.0366$ \\
KNN-QSMOTE       & $0.7267 \pm 0.0198$ & $0.7123 \pm 0.0942$ & $0.1133 \pm 0.0419$ & $0.1917 \pm 0.0633$ \\
Fidelity-QSMOTE  & $0.7485 \pm 0.0244$ & $0.7361 \pm 0.0458$ & $0.5886 \pm 0.0697$ & $0.6503 \pm 0.0382$ \\
Margin-QSMOTE    & $0.7936 \pm 0.0223$ & $0.8355 \pm 0.0319$ & $0.6650 \pm 0.0375$ & $0.7398 \pm 0.0287$ \\
\hline
\multicolumn{5}{c}{$\mathbf{n\_copies=2}$} \\
\hline
\textbf{QSMOTE Variant} & \textbf{Accuracy} & \textbf{Precision} & \textbf{Recall} & \textbf{F1 Score} \\
\hline
QSMOTE           & $0.8074 \pm 0.0099$ & $0.7817 \pm 0.0144$ & $0.8594 \pm 0.0168$ & $0.8185 \pm 0.0069$ \\
KNN-QSMOTE       & $0.7711 \pm 0.0180$ & $0.6661 \pm 0.0351$ & $0.4431 \pm 0.0726$ & $0.5273 \pm 0.0487$ \\
Fidelity-QSMOTE  & $0.7875 \pm 0.0152$ & $0.7271 \pm 0.0333$ & $0.7540 \pm 0.0236$ & $0.7395 \pm 0.0155$ \\
Margin-QSMOTE    & $0.8398 \pm 0.0160$ & $0.8375 \pm 0.0220$ & $0.7859 \pm 0.0347$ & $0.8102 \pm 0.0197$ \\
\hline
\end{tabular}
\label{tab:qsmote_pgm_vertical}
\end{table*}

Table~\ref{tab:qsmote_rf} summarizes the performance of QSMOTE and its variants when combined with a RF classifier. The baseline QSMOTE method achieved balanced results with an accuracy of $0.7995 \pm 0.0187$ and F1-score of $0.7457 \pm 0.0268$. The KNN-based variant performed the weakest overall, with lower recall ($0.5224 \pm 0.0584$) and F1-score ($0.5743 \pm 0.0324$), suggesting that neighborhood-based resampling may not be well-suited for this dataset. In contrast, the Fidelity-based approach provided a notable improvement in recall ($0.7940 \pm 0.0300$) and F1-score ($0.7785 \pm 0.0298$), indicating more effective minority class synthesis. The Margin-filtered QSMOTE variant obtained the highest accuracy ($0.8371 \pm 0.0226$) and precision ($0.8073 \pm 0.0412$), though with a slight trade-off in recall compared to the Fidelity-based method. Overall, these results highlight the advantage of Fidelity- and Margin-based strategies in generating synthetic samples that improve classifier performance compared to standard and KNN-based QSMOTE.
Table~\ref{tab:qsmote_pgm_vertical} presents the comparative performance of QSMOTE and its variants under the PGM classifier with amplitude encoding for $n\_copies=1$ and $n\_copies=2$. For $n\_copies=1$, Margin-QSMOTE achieved the best overall balance with the highest accuracy ($0.7936 \pm 0.0223$) and F1-score ($0.7398 \pm 0.0287$), while QSMOTE also performed competitively with strong precision but lower recall. In contrast, KNN-QSMOTE suffered from very low recall ($0.1133 \pm 0.0419$), leading to poor F1 performance. With $n\_copies=2$, all variants improved, with QSMOTE showing a substantial gain in recall ($0.8594 \pm 0.0168$) and F1-score ($0.8185 \pm 0.0069$). Margin-QSMOTE remained the top performer overall, achieving the highest accuracy ($0.8398 \pm 0.0160$) and strong balanced metrics, while Fidelity-QSMOTE provided stable improvements across all metrics. These results indicate that increasing the number of copies enhances classifier performance, with Margin- and Fidelity-based resampling strategies yielding the most consistent benefits.

\begin{table*}[htbp]
\centering
\caption{Performance of QSMOTE and its variants with the PGM classifier using stereo encoding for $n\_copies=1,2$.}
\begin{tabular}{lcccc}
\hline
\multicolumn{5}{c}{$\mathbf{n\_copies=1}$} \\
\hline
\textbf{QSMOTE Variant} & \textbf{Accuracy} & \textbf{Precision} & \textbf{Recall} & \textbf{F1 Score} \\
\hline
QSMOTE           & $0.7893 \pm 0.0322$ & $0.8315 \pm 0.0305$ & $0.6707 \pm 0.0587$ & $0.7407 \pm 0.0361$ \\
KNN-QSMOTE       & $0.7413 \pm 0.0215$ & $0.7426 \pm 0.0954$ & $0.1821 \pm 0.0522$ & $0.2876 \pm 0.0693$ \\
Fidelity-QSMOTE  & $0.7483 \pm 0.0257$ & $0.7331 \pm 0.0469$ & $0.5928 \pm 0.0724$ & $0.6516 \pm 0.0402$ \\
Margin-QSMOTE    & $0.7902 \pm 0.0206$ & $0.8388 \pm 0.0321$ & $0.6508 \pm 0.0436$ & $0.7313 \pm 0.0235$ \\
\hline
\multicolumn{5}{c}{$\mathbf{n\_copies=2}$} \\
\hline
\textbf{QSMOTE Variant} & \textbf{Accuracy} & \textbf{Precision} & \textbf{Recall} & \textbf{F1 Score} \\
\hline
QSMOTE           & $0.8512 \pm 0.0181$ & $0.8652 \pm 0.0328$ & $0.7865 \pm 0.0291$ & $0.8234 \pm 0.0235$ \\
KNN-QSMOTE       & $0.7733 \pm 0.0152$ & $0.6697 \pm 0.0381$ & $0.4489 \pm 0.0562$ & $0.5343 \pm 0.0381$ \\
Fidelity-QSMOTE  & $0.7801 \pm 0.0167$ & $0.7241 \pm 0.0353$ & $0.7318 \pm 0.0327$ & $0.7269 \pm 0.0183$ \\
Margin-QSMOTE    & $0.8341 \pm 0.0236$ & $0.8545 \pm 0.0260$ & $0.7628 \pm 0.0376$ & $0.8053 \pm 0.0231$ \\
\hline
\end{tabular}
\label{tab:qsmote_pgm_stereo}
\end{table*}

\begin{table*}[htbp]
\centering
\caption{Performance of QSMOTE and its variants with the KPGM classifier using amplitude encoding for $n\_copies=1,2$.}
\begin{tabular}{lcccc}
\hline
\multicolumn{5}{c}{$\mathbf{n\_copies=1}$} \\
\hline
\textbf{QSMOTE Variant} & \textbf{Accuracy} & \textbf{Precision} & \textbf{Recall} & \textbf{F1 Score} \\
\hline
QSMOTE           & $0.7878 \pm 0.0168$ & $0.8229 \pm 0.0181$ & $0.6655 \pm 0.0358$ & $0.7350 \pm 0.0180$ \\
KNN-QSMOTE       & $0.7295 \pm 0.0204$ & $0.7043 \pm 0.0887$ & $0.1306 \pm 0.0528$ & $0.2153 \pm 0.0780$ \\
Fidelity-QSMOTE  & $0.7483 \pm 0.0249$ & $0.7359 \pm 0.0463$ & $0.5880 \pm 0.0691$ & $0.6499 \pm 0.0384$ \\
Margin-QSMOTE    & $0.7801 \pm 0.0197$ & $0.8273 \pm 0.0391$ & $0.6231 \pm 0.0332$ & $0.7098 \pm 0.0230$ \\
\hline
\multicolumn{5}{c}{$\mathbf{n\_copies=2}$} \\
\hline
\textbf{QSMOTE Variant} & \textbf{Accuracy} & \textbf{Precision} & \textbf{Recall} & \textbf{F1 Score} \\
\hline
QSMOTE           & $0.8389 \pm 0.0178$ & $0.8405 \pm 0.0358$ & $0.7763 \pm 0.0337$ & $0.8061 \pm 0.0205$ \\
KNN-QSMOTE       & $0.7733 \pm 0.0190$ & $0.6678 \pm 0.0396$ & $0.4552 \pm 0.0660$ & $0.5373 \pm 0.0439$ \\
Fidelity-QSMOTE  & $0.7862 \pm 0.0149$ & $0.7256 \pm 0.0342$ & $0.7519 \pm 0.0227$ & $0.7377 \pm 0.0163$ \\
Margin-QSMOTE    & $0.8483 \pm 0.0147$ & $0.8426 \pm 0.0365$ & $0.8057 \pm 0.0257$ & $0.8230 \pm 0.0197$ \\
\hline
\end{tabular}
\label{tab:qsmote_kpgm_amp}
\end{table*}
Table~\ref{tab:qsmote_pgm_stereo} reports the results of QSMOTE and its variants under the PGM classifier with stereo encoding. For $n\_copies=1$, QSMOTE and Margin-QSMOTE achieved comparable accuracy (around $0.79$) and F1-scores ($0.7407$ and $0.7313$ respectively), with QSMOTE showing higher recall while Margin-QSMOTE yielded stronger precision. Fidelity-QSMOTE provided moderate but consistent results, whereas KNN-QSMOTE again underperformed due to very low recall, leading to the weakest F1-score ($0.2876$). When the number of copies was increased to $n\_copies=2$, overall performance improved substantially. QSMOTE recorded the highest accuracy ($0.8512 \pm 0.0181$) and F1-score ($0.8234 \pm 0.0235$), demonstrating strong recall gains, while Margin-QSMOTE also performed competitively with balanced precision and recall. Fidelity-QSMOTE remained stable, though with lower accuracy, and KNN-QSMOTE continued to lag behind. These results confirm that stereo encoding with additional copies enhances the effectiveness of the PGM classifier, with QSMOTE and Margin-QSMOTE emerging as the most reliable approaches.
Table~\ref{tab:qsmote_kpgm_amp} presents the results of QSMOTE and its variants under the KPGM classifier with amplitude encoding. For $n\_copies=1$, QSMOTE achieved solid performance with accuracy of $0.7878 \pm 0.0168$ and F1-score of $0.7350 \pm 0.0180$, while Margin-QSMOTE yielded similar accuracy ($0.7801 \pm 0.0197$) and slightly stronger precision but lower recall. Fidelity-QSMOTE provided moderate but balanced improvements, whereas KNN-QSMOTE performed poorly, with recall dropping to only $0.1306 \pm 0.0528$, resulting in the weakest F1-score. With $n\_copies=2$, all methods improved considerably, with Margin-QSMOTE emerging as the best-performing variant, achieving the highest accuracy ($0.8483 \pm 0.0147$) and F1-score ($0.8230 \pm 0.0197$). QSMOTE also performed strongly, with accuracy of $0.8389 \pm 0.0178$ and F1-score of $0.8061 \pm 0.0205$, showing significant gains in recall. Fidelity-QSMOTE remained consistent but slightly lower-performing, while KNN-QSMOTE, although improved, continued to underperform compared to other variants. These findings confirm that increasing the number of copies enhances the classification ability of KPGM, with Margin-QSMOTE and QSMOTE standing out as the most effective resampling strategies.

\begin{table*}[htbp]
\centering
\caption{Performance of QSMOTE and its variants with the KPGM classifier using stereo encoding for $n\_copies=1,2$.}
\begin{tabular}{lcccc}
\hline
\multicolumn{5}{c}{$\mathbf{n\_copies=1}$} \\
\hline
\textbf{QSMOTE Variant} & \textbf{Accuracy} & \textbf{Precision} & \textbf{Recall} & \textbf{F1 Score} \\
\hline
QSMOTE           & $0.8078 \pm 0.0176$ & $0.8409 \pm 0.0384$ & $0.6951 \pm 0.0302$ & $0.7602 \pm 0.0218$ \\
KNN-QSMOTE       & $0.7337 \pm 0.0195$ & $0.7055 \pm 0.0781$ & $0.1603 \pm 0.0475$ & $0.2565 \pm 0.0646$ \\
Fidelity-QSMOTE  & $0.7483 \pm 0.0248$ & $0.7332 \pm 0.0466$ & $0.5928 \pm 0.0706$ & $0.6517 \pm 0.0384$ \\
Margin-QSMOTE    & $0.7894 \pm 0.0191$ & $0.8359 \pm 0.0362$ & $0.6521 \pm 0.0399$ & $0.7310 \pm 0.0174$ \\
\hline
\multicolumn{5}{c}{$\mathbf{n\_copies=2}$} \\
\hline
\textbf{QSMOTE Variant} & \textbf{Accuracy} & \textbf{Precision} & \textbf{Recall} & \textbf{F1 Score} \\
\hline
QSMOTE           & $0.8511 \pm 0.0165$ & $0.8629 \pm 0.0289$ & $0.7868 \pm 0.0286$ & $0.8225 \pm 0.0189$ \\
KNN-QSMOTE       & $0.7702 \pm 0.0145$ & $0.6646 \pm 0.0406$ & $0.4378 \pm 0.0509$ & $0.5248 \pm 0.0351$ \\
Fidelity-QSMOTE  & $0.7806 \pm 0.0160$ & $0.7248 \pm 0.0353$ & $0.7324 \pm 0.0328$ & $0.7274 \pm 0.0181$ \\
Margin-QSMOTE    & $0.8445 \pm 0.0197$ & $0.8600 \pm 0.0347$ & $0.7658 \pm 0.0309$ & $0.8094 \pm 0.0204$ \\
\hline
\end{tabular}
\label{tab:qsmote_kpgm_stereo}
\end{table*}

\begin{figure*}[htbp]
  \centering
  \begin{subfigure}{0.48\textwidth}
    \centering
    \includegraphics[width=\linewidth]{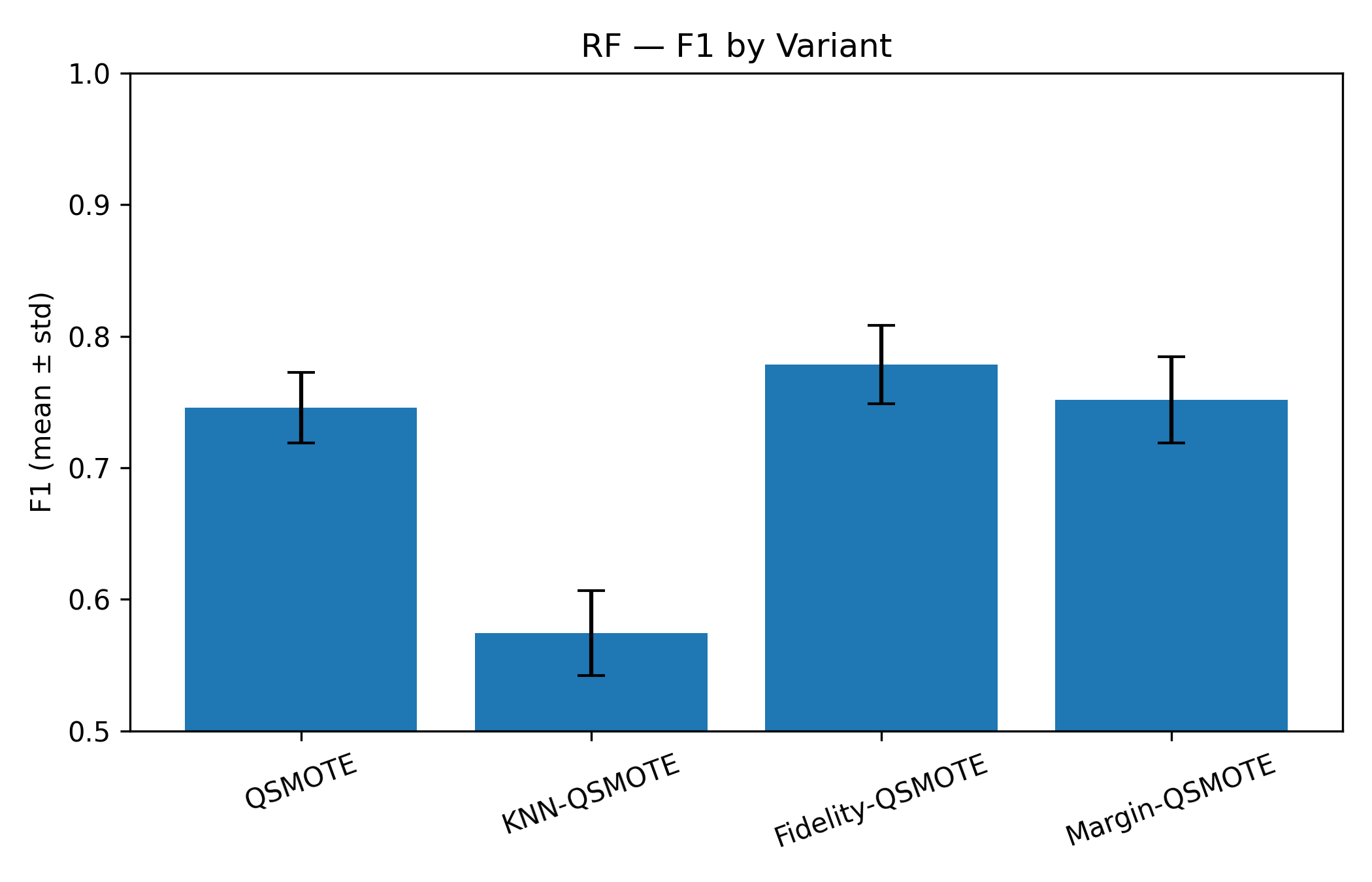}
    \caption{RF: F1 (mean $\pm$ std) across QSMOTE variants}
    \label{fig:rf_f1}
  \end{subfigure}\hfill
  \begin{subfigure}{0.48\textwidth}
    \centering
    \includegraphics[width=0.8\linewidth]{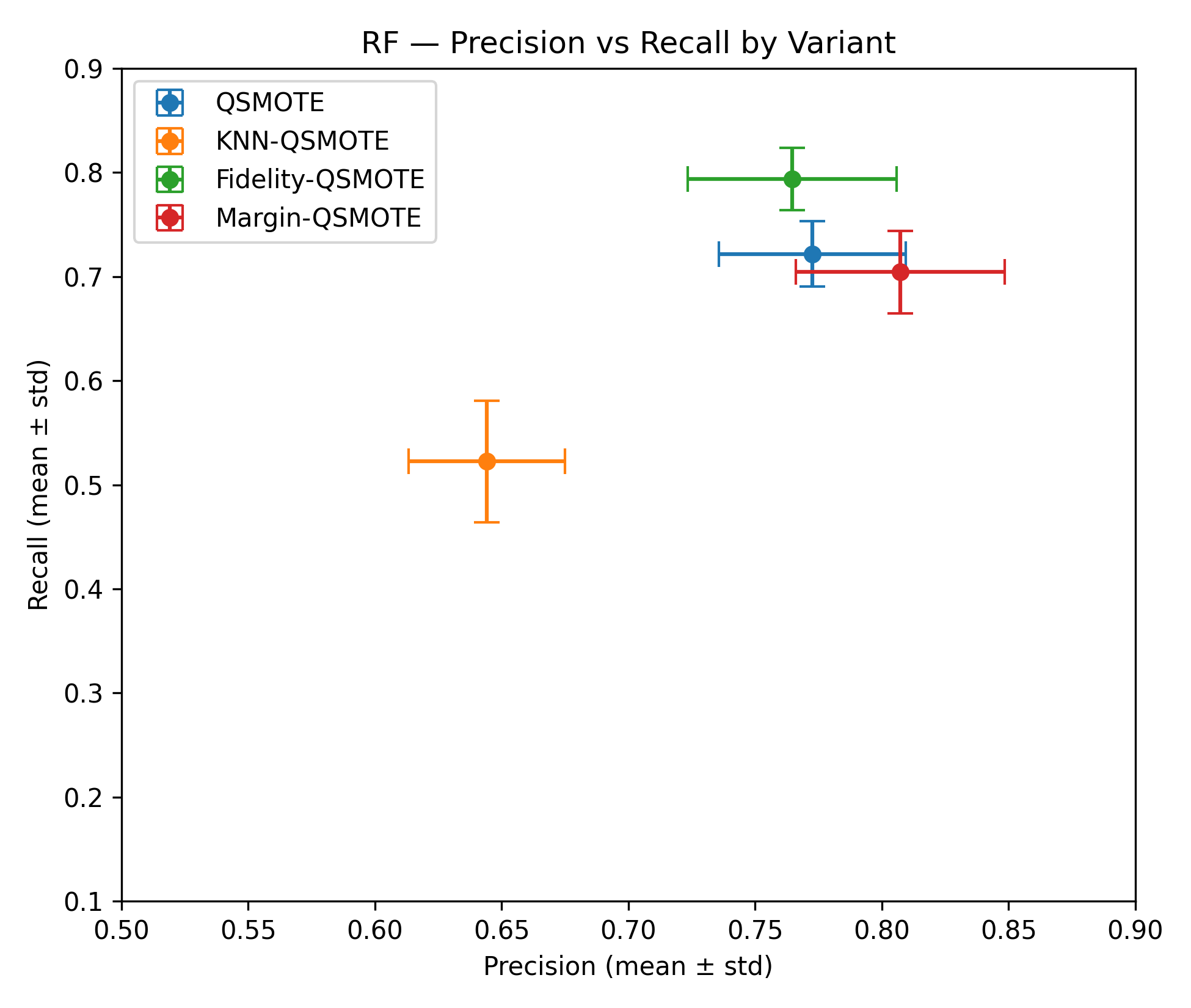}
    \caption{RF: Precision-Recall tradeoff (mean $\pm$ std)}
    \label{fig:rf_pr}
  \end{subfigure}
  \caption{RF baseline performance across QSMOTE variants.}
  \label{fig:rf_overview}
\end{figure*}

\begin{figure*}[htbp]
  \centering
  \begin{subfigure}{0.48\textwidth}
    \centering
    \includegraphics[width=\linewidth]{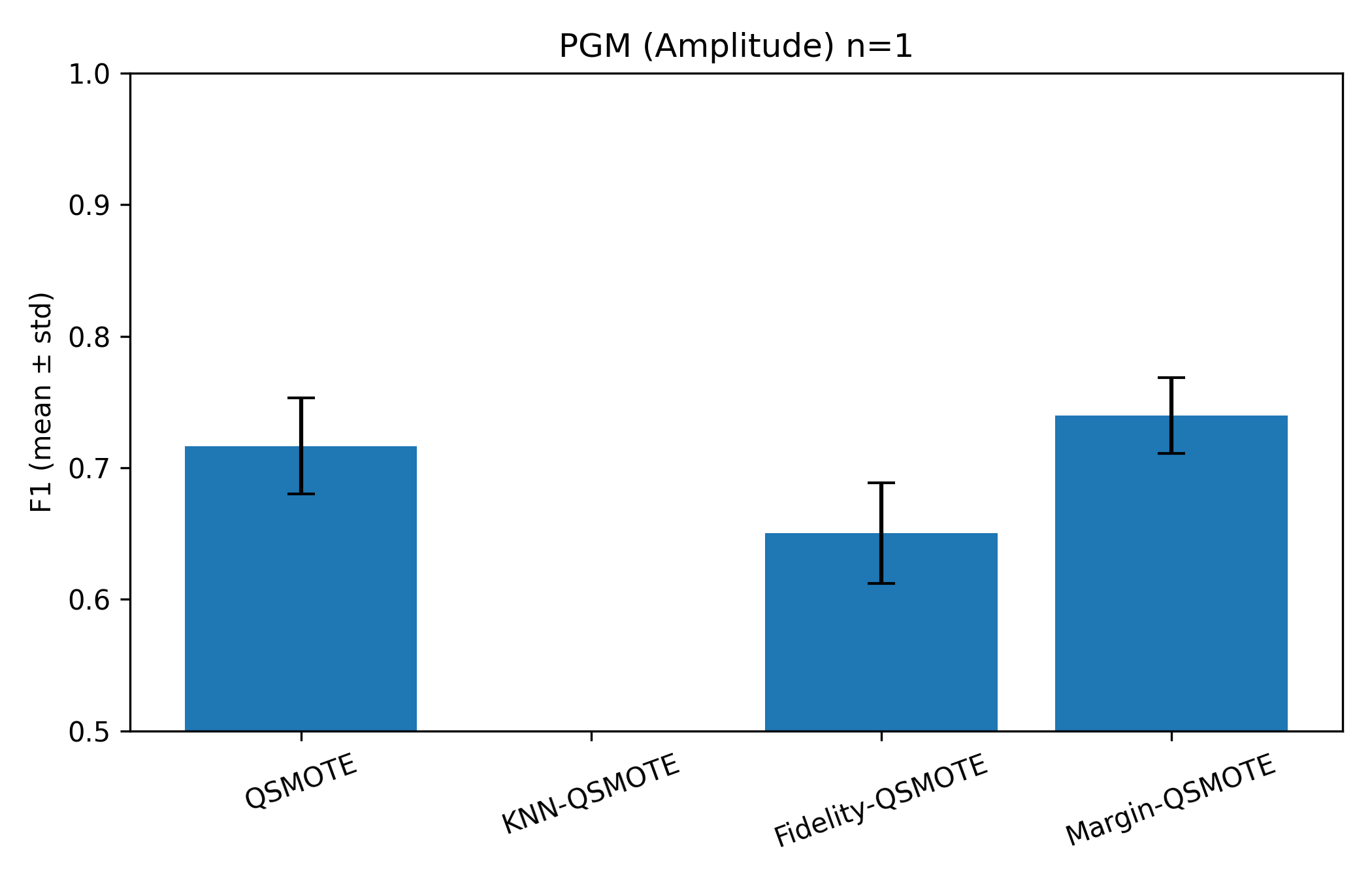}
    \caption{PGM (Amplitude), $n\_copies=1$}
    \label{fig:pgm_amp_n1_f1}
  \end{subfigure}\hfill
  \begin{subfigure}{0.48\textwidth}
    \centering
    \includegraphics[width=\linewidth]{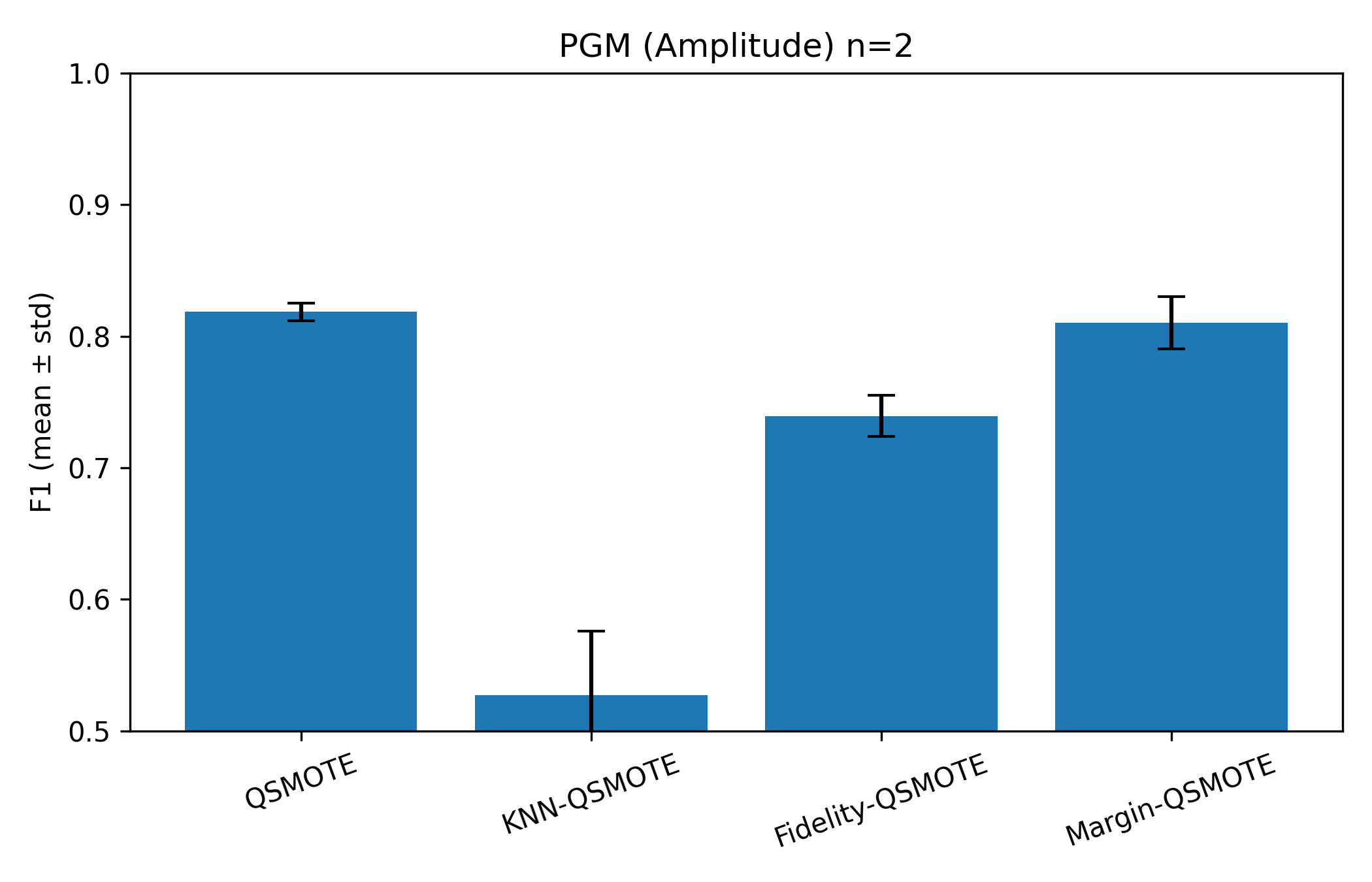}
    \caption{PGM (Amplitude), $n\_copies=2$}
    \label{fig:pgm_amp_n2_f1}
  \end{subfigure}
  \caption{F1 (mean $\pm$ std) by QSMOTE variant for PGM with amplitude encoding.}
  \label{fig:pgm_amp_f1_bars}
\end{figure*}

\begin{figure*}[htbp]
  \centering
  \begin{subfigure}{0.48\textwidth}
    \centering
    \includegraphics[width=\linewidth]{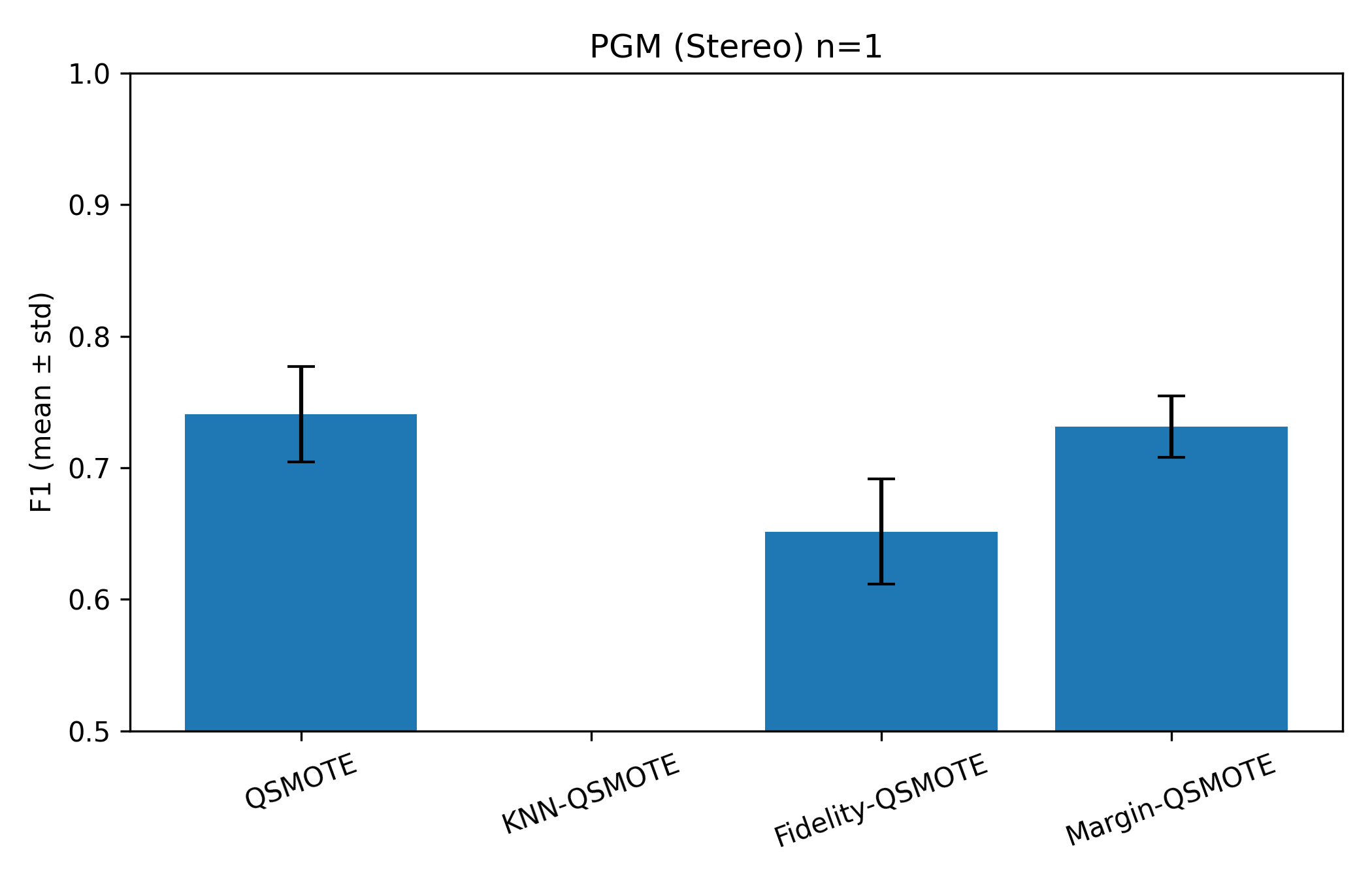}
    \caption{PGM (Stereo), $n\_copies=1$}
    \label{fig:pgm_st_n1_f1}
  \end{subfigure}\hfill
  \begin{subfigure}{0.48\textwidth}
    \centering
    \includegraphics[width=\linewidth]{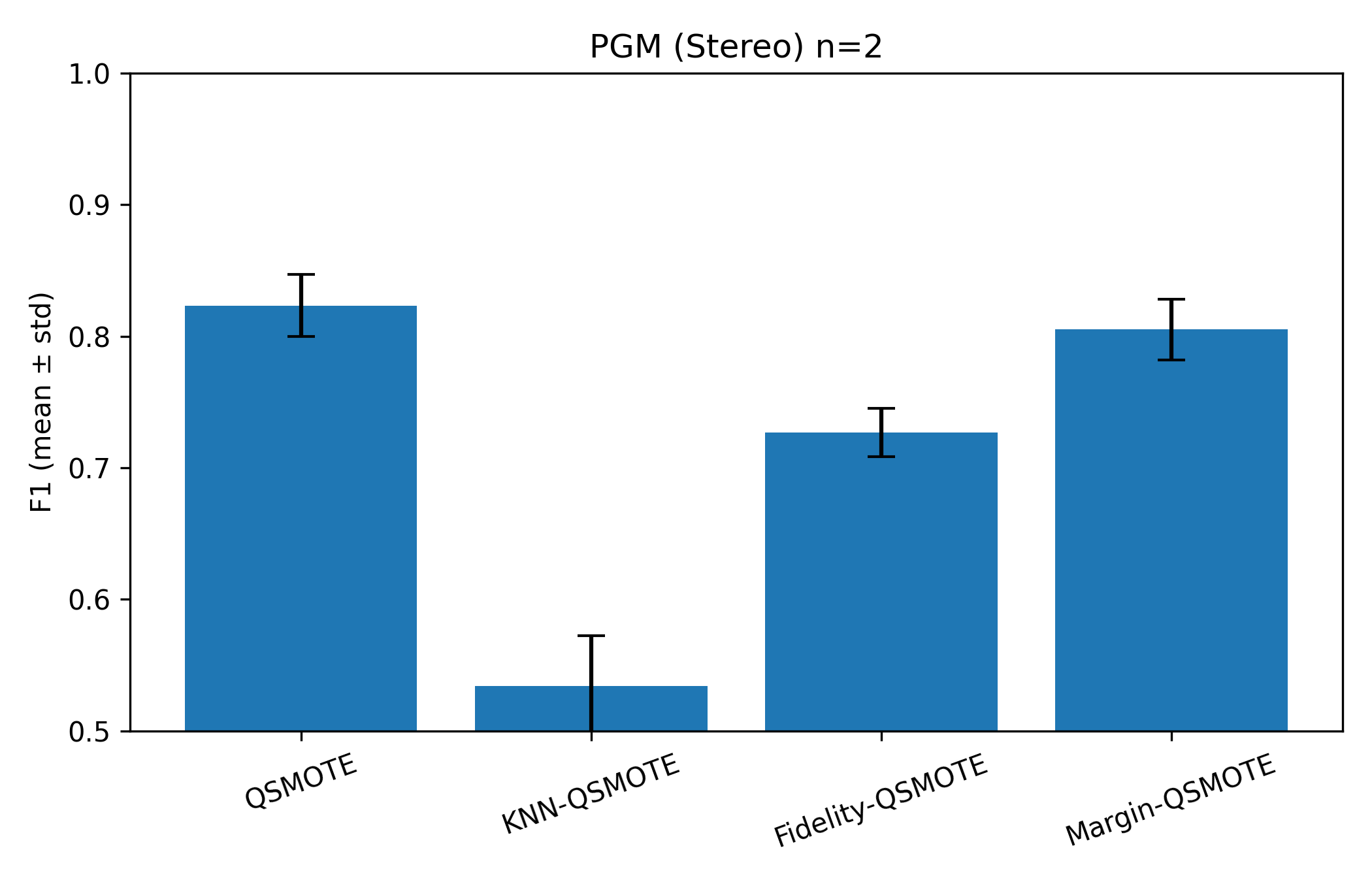}
    \caption{PGM (Stereo), $n\_copies=2$}
    \label{fig:pgm_st_n2_f1}
  \end{subfigure}
  \caption{F1 (mean $\pm$ std) by QSMOTE variant for PGM with stereo encoding.}
  \label{fig:pgm_stereo_f1_bars}
\end{figure*}

\begin{figure*}[htbp]
  \centering
  \begin{subfigure}{0.48\textwidth}
    \centering
    \includegraphics[width=\linewidth]{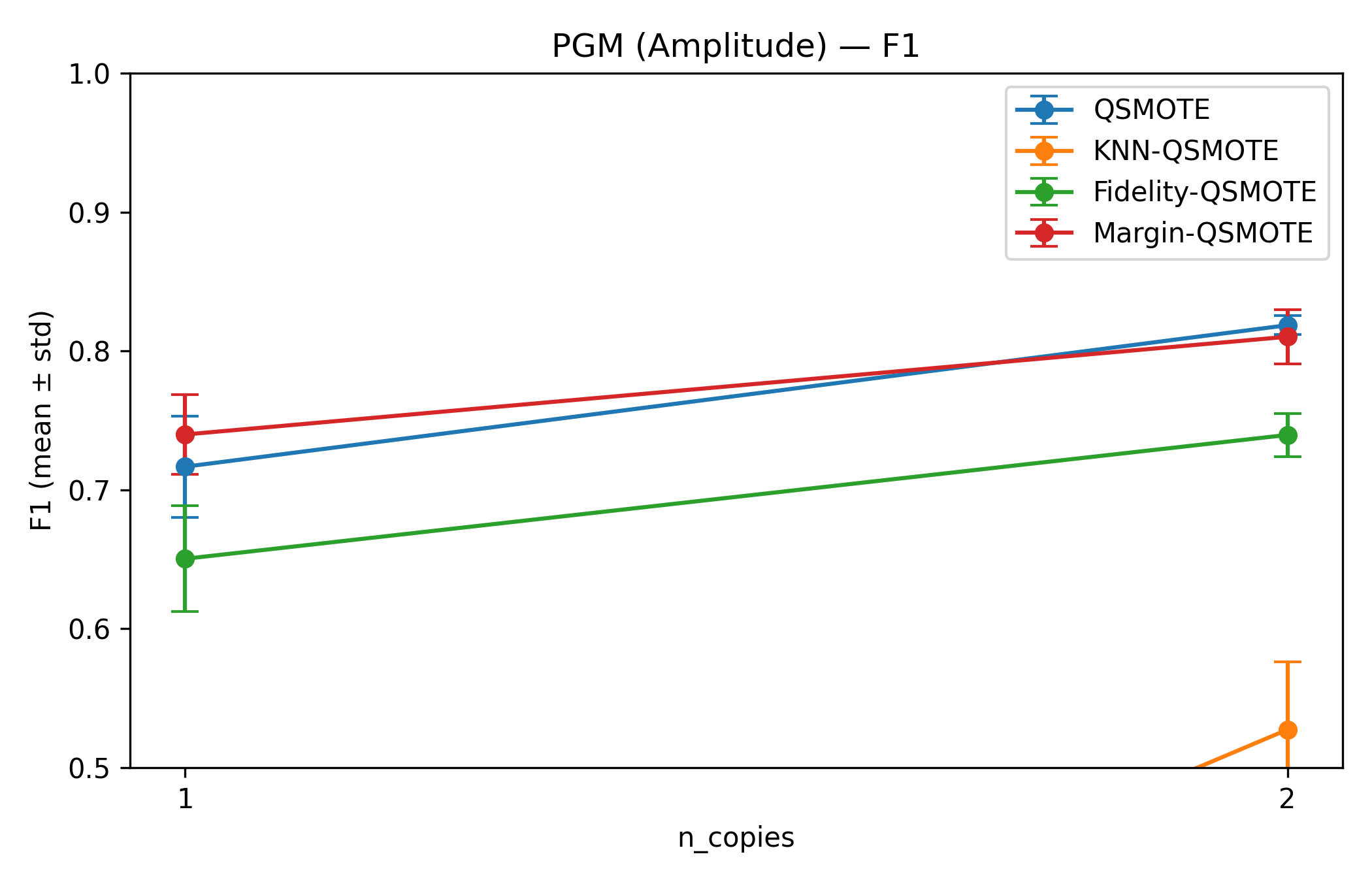}
    \caption{PGM (Amplitude): F1 vs $n\_copies$}
    \label{fig:pgm_amp_effect_copies}
  \end{subfigure}\hfill
  \begin{subfigure}{0.48\textwidth}
    \centering
    \includegraphics[width=\linewidth]{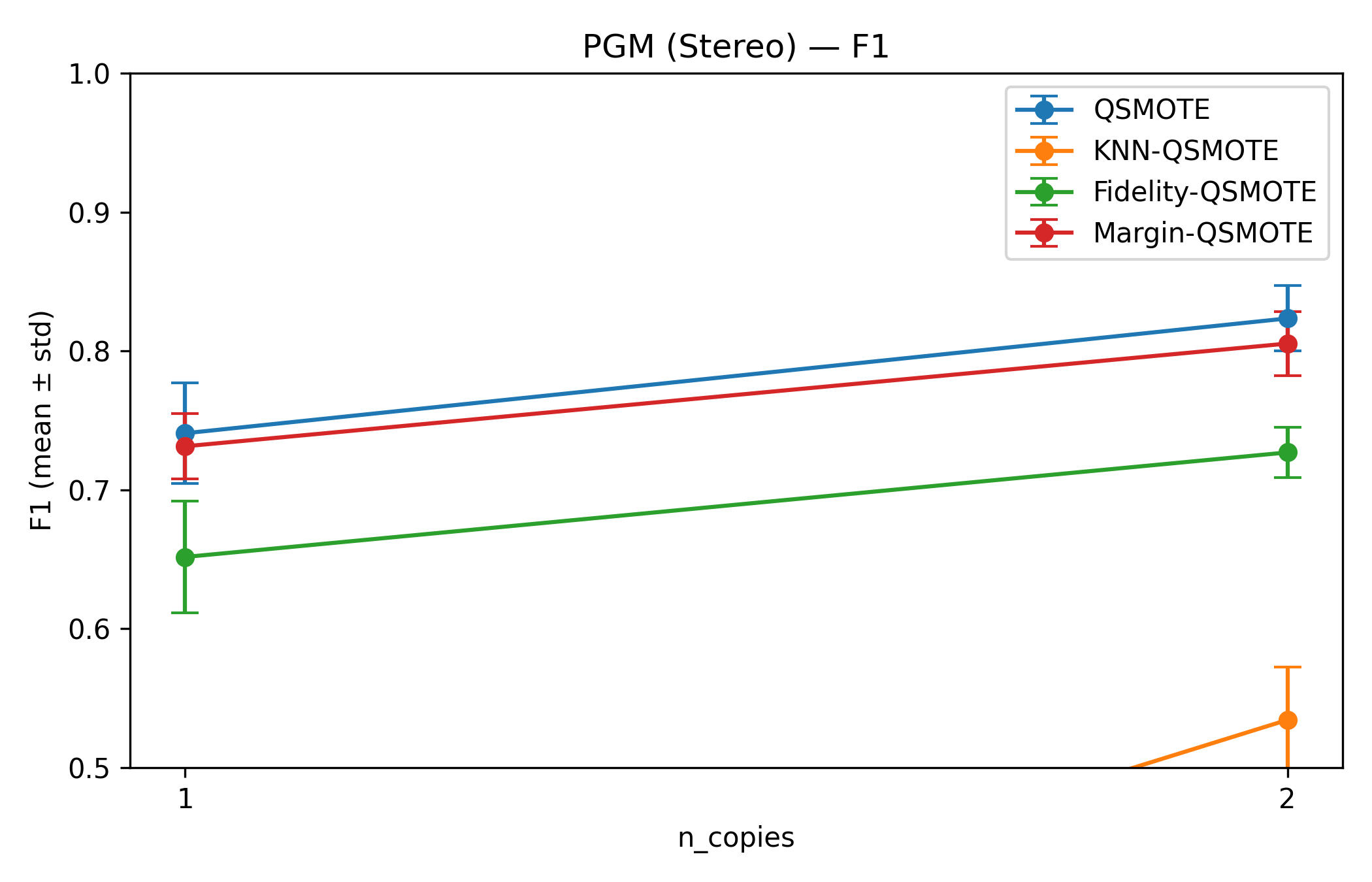}
    \caption{PGM (Stereo): F1 vs $n\_copies$}
    \label{fig:pgm_stereo_effect_copies}
  \end{subfigure}
  \caption{Effect of $n\_copies$ on PGM performance across QSMOTE variants.}
  \label{fig:pgm_effect_copies}
\end{figure*}

\begin{figure*}[htbp]
  \centering
  \begin{subfigure}{0.48\textwidth}
    \centering
    \includegraphics[width=\linewidth]{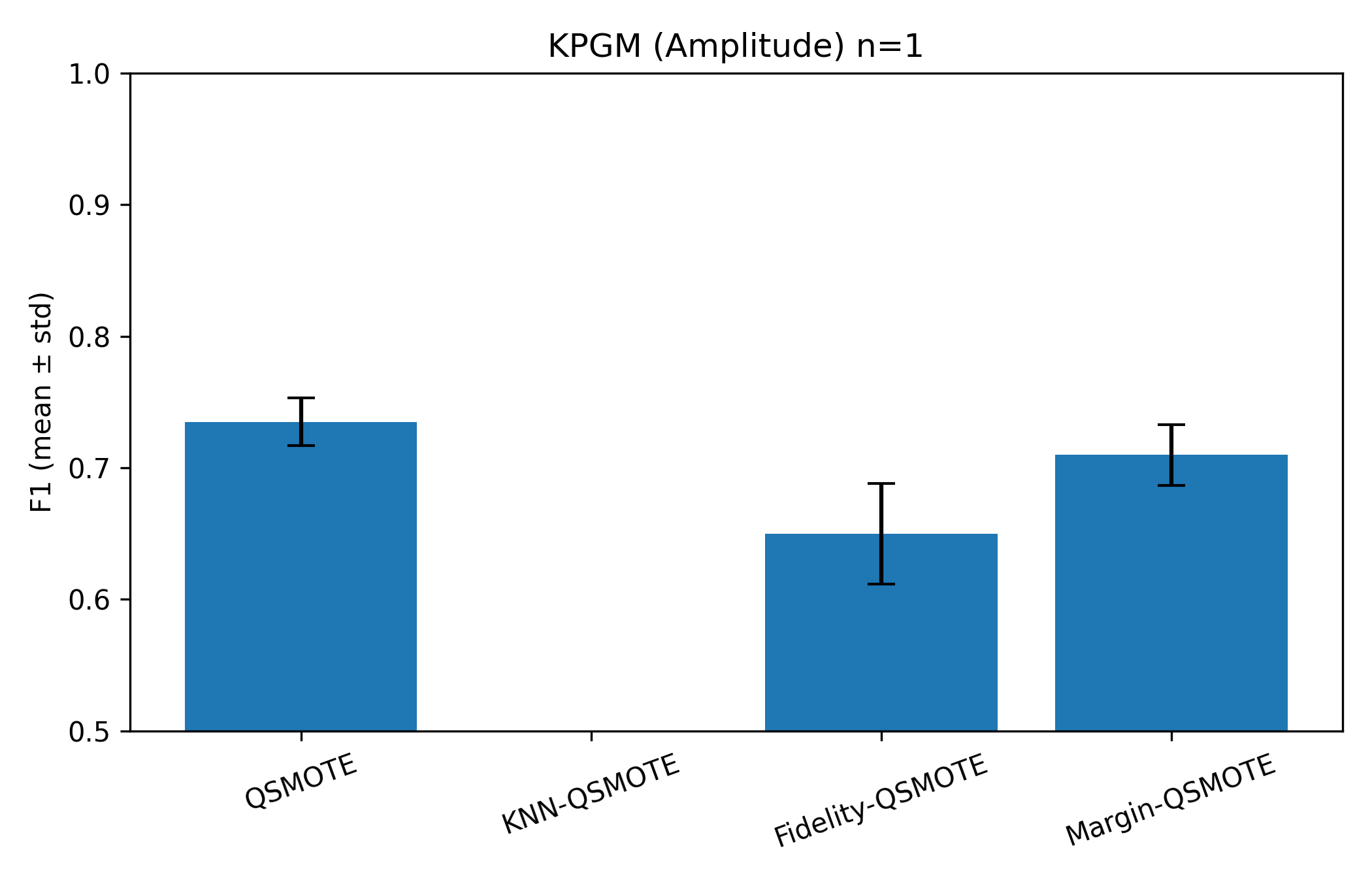}
    \caption{KPGM (Amplitude), $n\_copies=1$}
    \label{fig:kpgm_amp_n1_f1}
  \end{subfigure}\hfill
  \begin{subfigure}{0.48\textwidth}
    \centering
    \includegraphics[width=\linewidth]{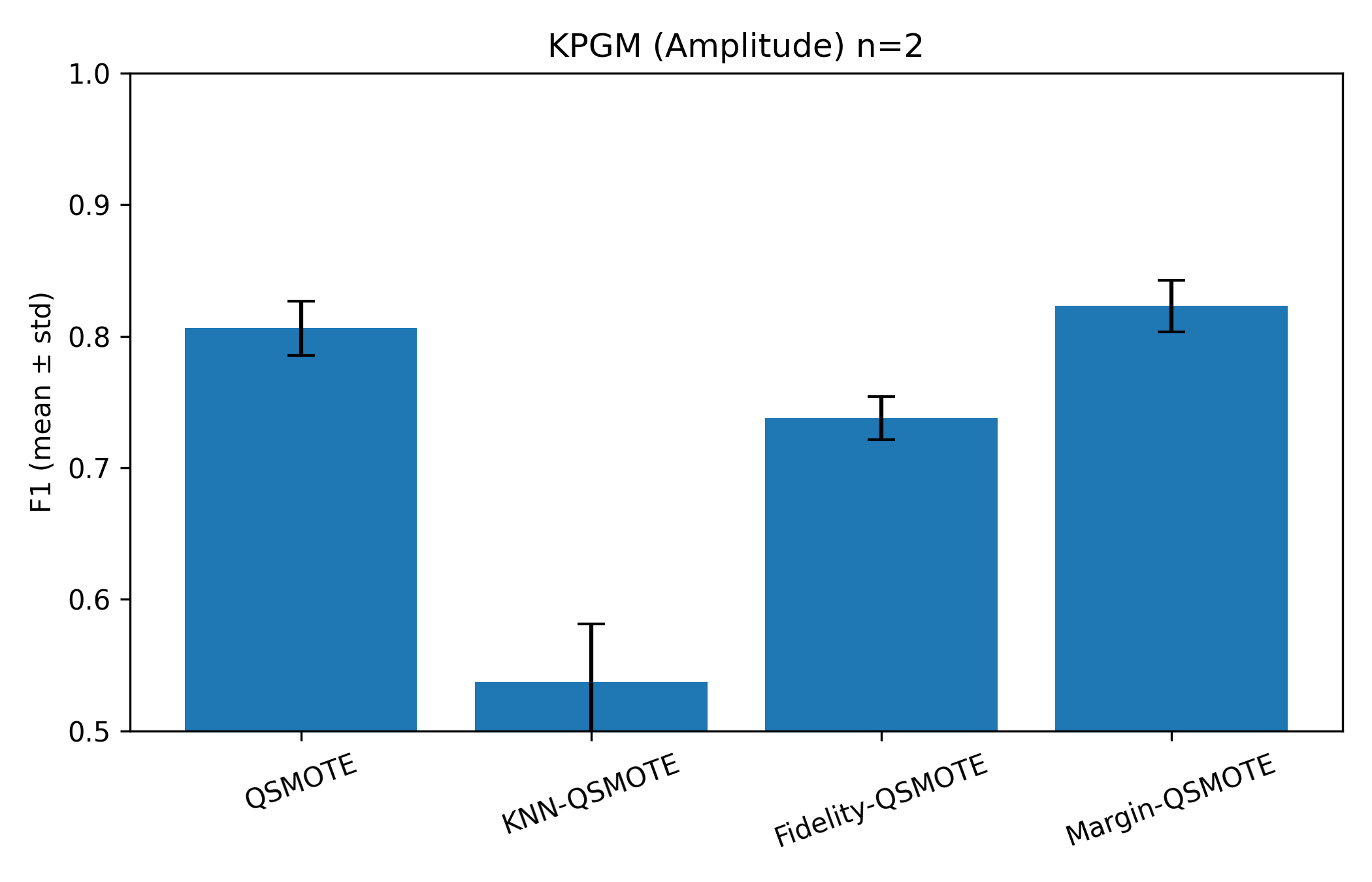}
    \caption{KPGM (Amplitude), $n\_copies=2$}
    \label{fig:kpgm_amp_n2_f1}
  \end{subfigure}
  \caption{F1 (mean $\pm$ std) by QSMOTE variant for KPGM with amplitude encoding.}
  \label{fig:kpgm_amp_f1_bars}
\end{figure*}

\begin{figure*}[htbp]
  \centering
  \begin{subfigure}{0.48\textwidth}
    \centering
    \includegraphics[width=\linewidth]{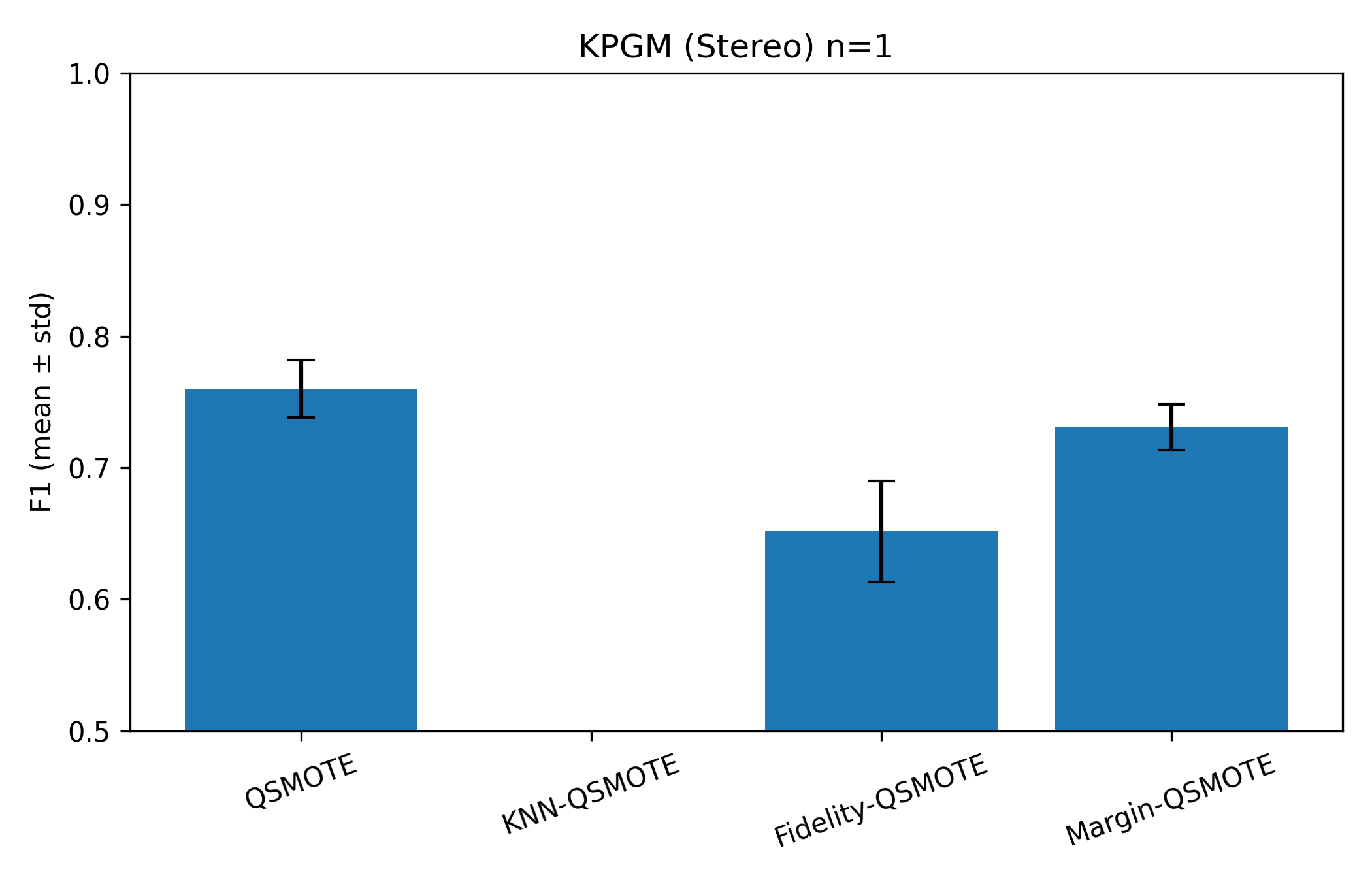}
    \caption{KPGM (Stereo), $n\_copies=1$}
    \label{fig:kpgm_st_n1_f1}
  \end{subfigure}\hfill
  \begin{subfigure}{0.48\textwidth}
    \centering
    \includegraphics[width=\linewidth]{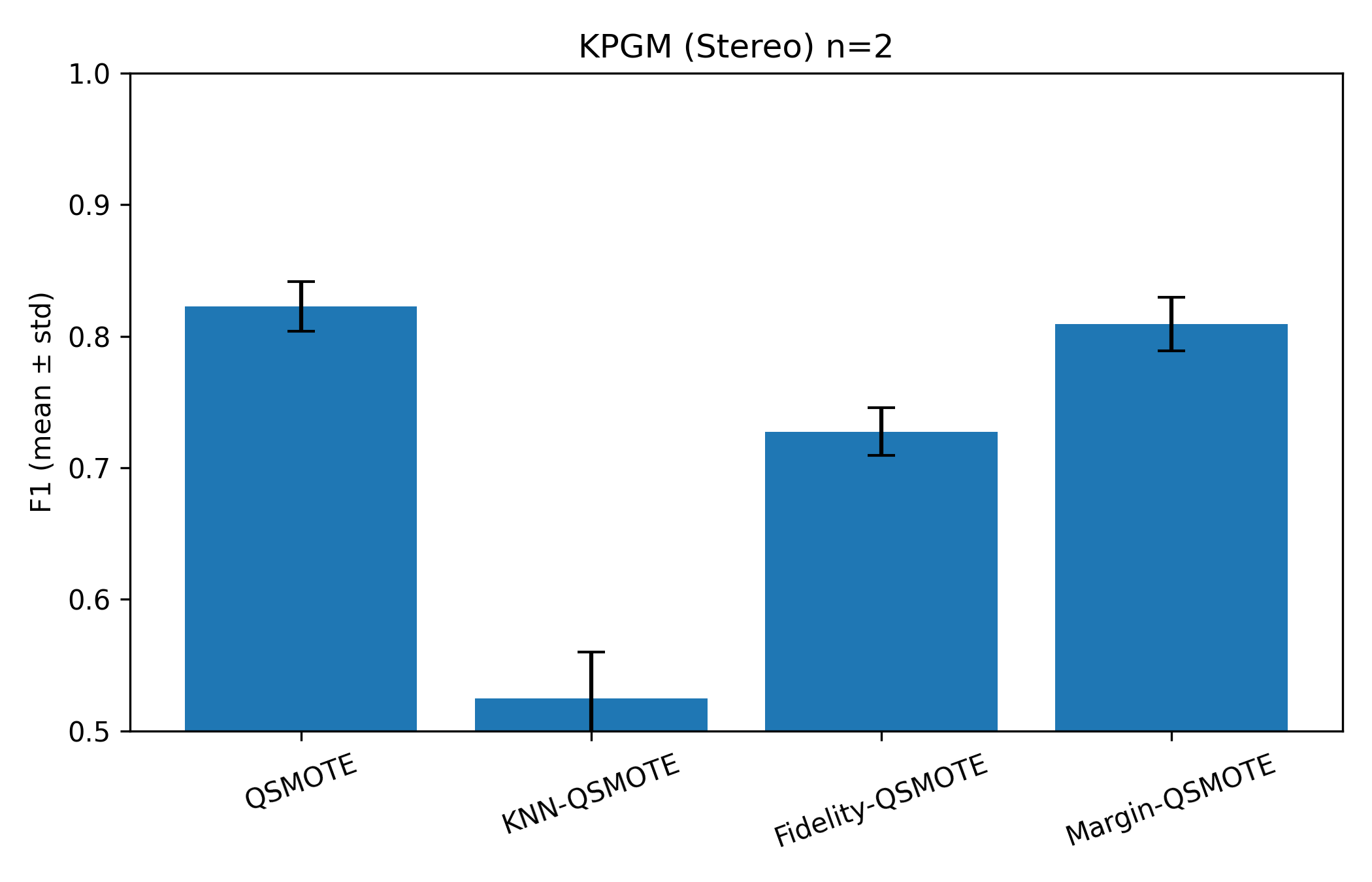}
    \caption{KPGM (Stereo), $n\_copies=2$}
    \label{fig:kpgm_st_n2_f1}
  \end{subfigure}
  \caption{F1 (mean $\pm$ std) by QSMOTE variant for KPGM with stereo encoding.}
  \label{fig:kpgm_stereo_f1_bars}
\end{figure*}

\begin{figure*}[htbp]
  \centering
  \begin{subfigure}{0.48\textwidth}
    \centering
    \includegraphics[width=\linewidth]{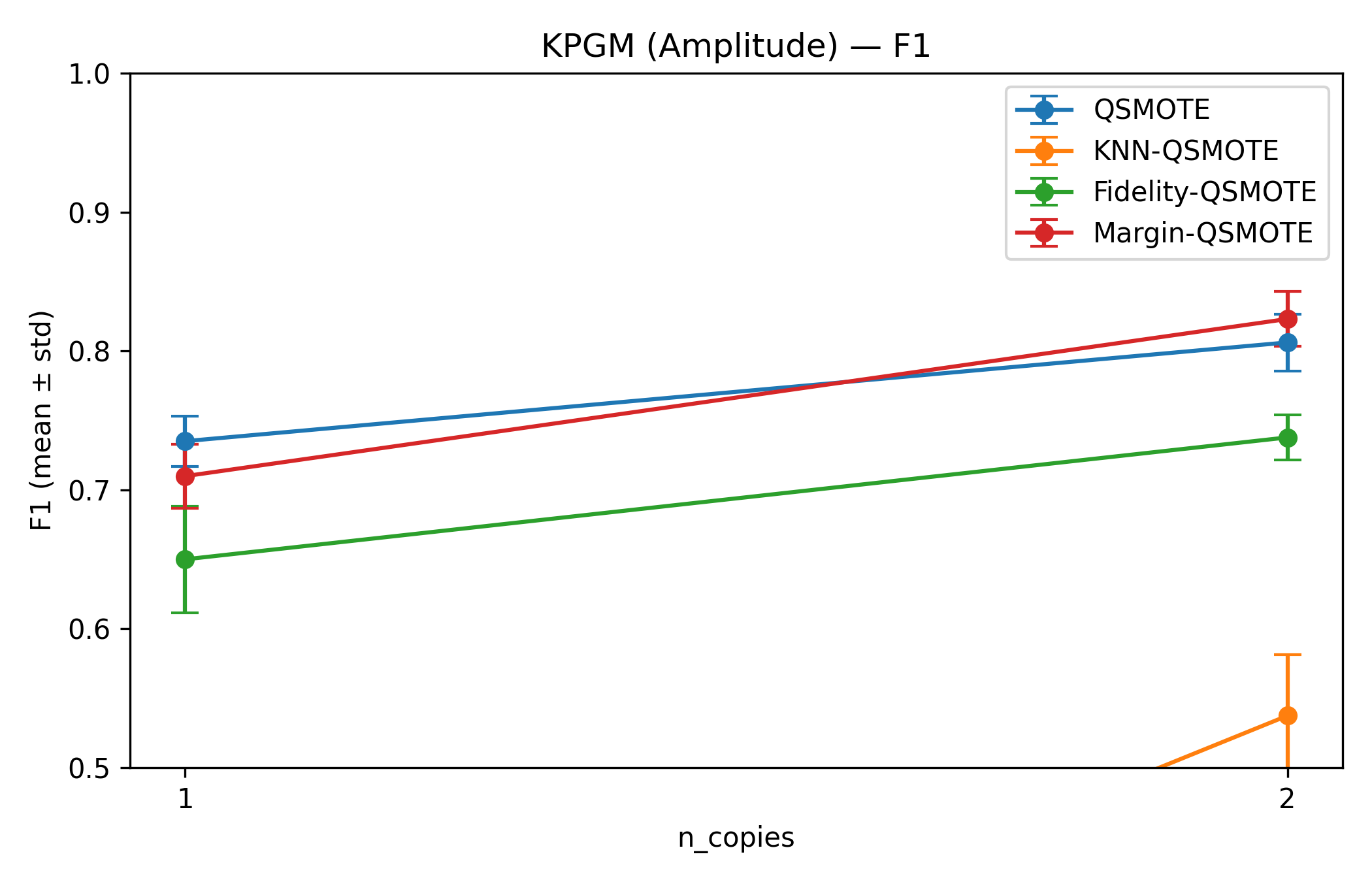}
    \caption{KPGM (Amplitude): F1 vs $n\_copies$}
    \label{fig:kpgm_amp_effect_copies}
  \end{subfigure}\hfill
  \begin{subfigure}{0.48\textwidth}
    \centering
    \includegraphics[width=\linewidth]{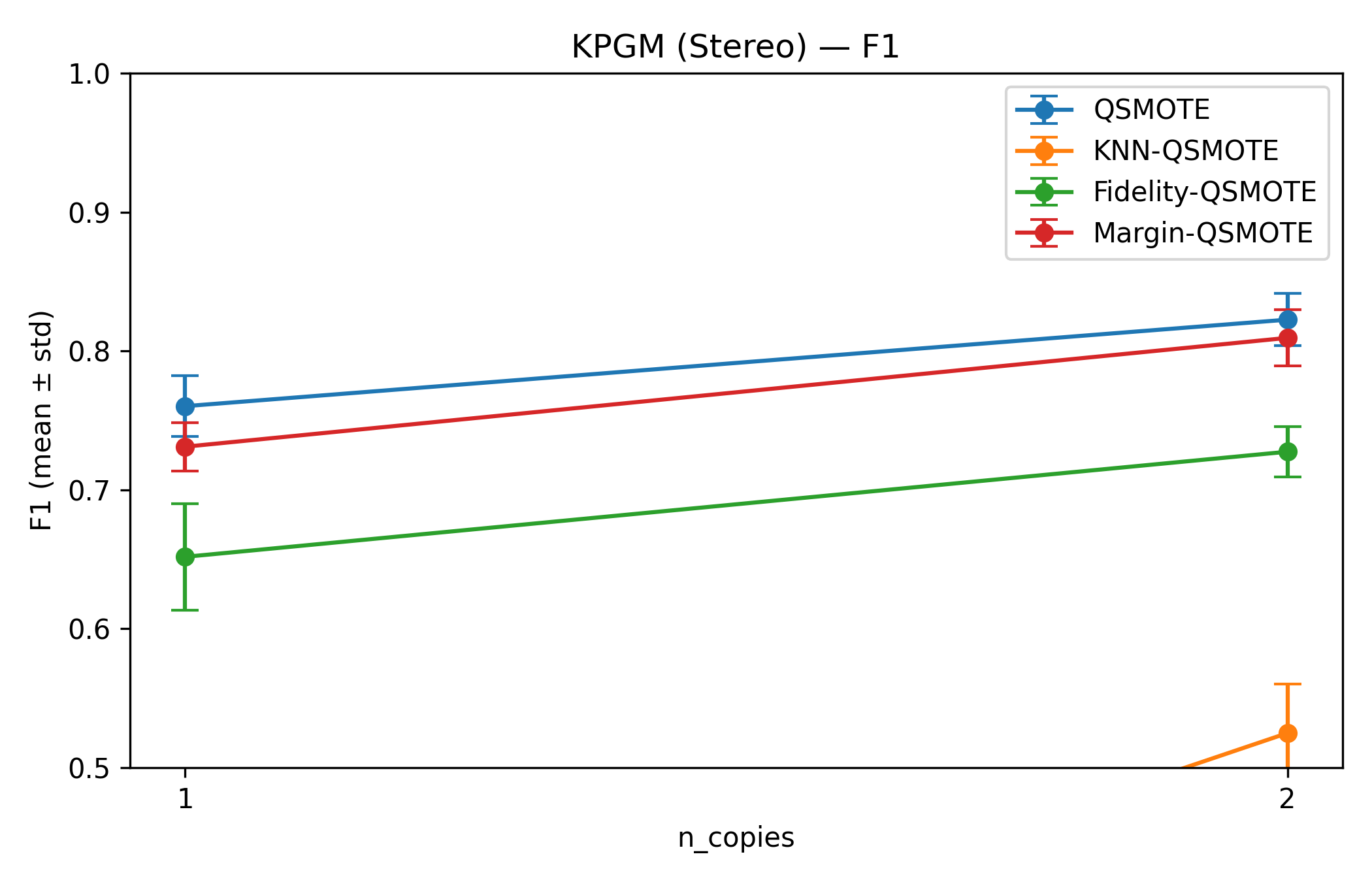}
    \caption{KPGM (Stereo): F1 vs $n\_copies$}
    \label{fig:kpgm_stereo_effect_copies}
  \end{subfigure}
  \caption{Effect of $n\_copies$ on KPGM performance across QSMOTE variants.}
  \label{fig:kpgm_effect_copies}
\end{figure*}

\begin{figure*}[htbp]
  \centering
  \begin{subfigure}{0.48\textwidth}
    \centering
    \includegraphics[width=0.8\linewidth]{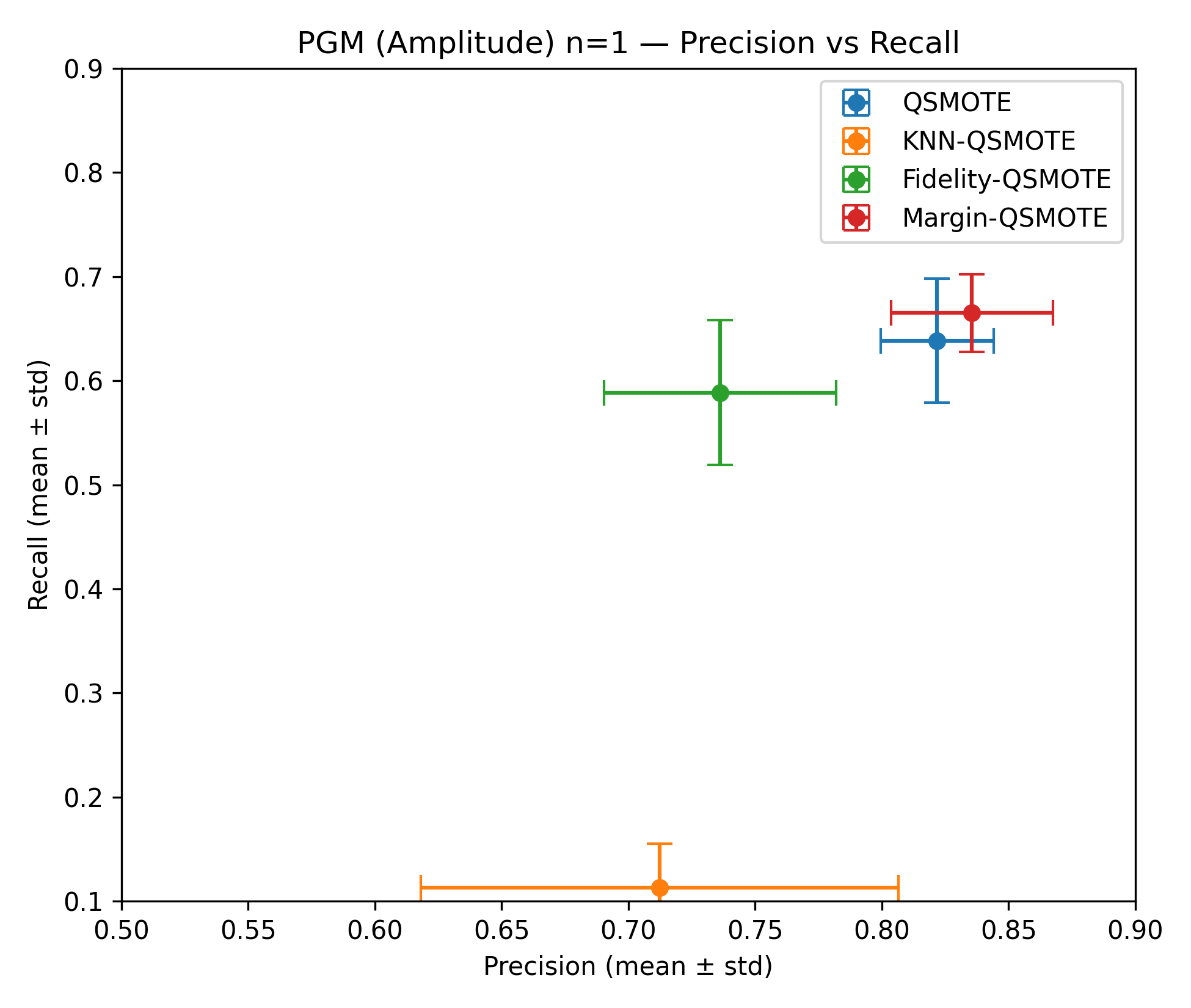}
    \caption{PGM (Amplitude), $n\_copies=1$}
    \label{fig:pgm_amp_n1_pr}
  \end{subfigure}\hfill
  \begin{subfigure}{0.48\textwidth}
    \centering
    \includegraphics[width=0.8\linewidth]{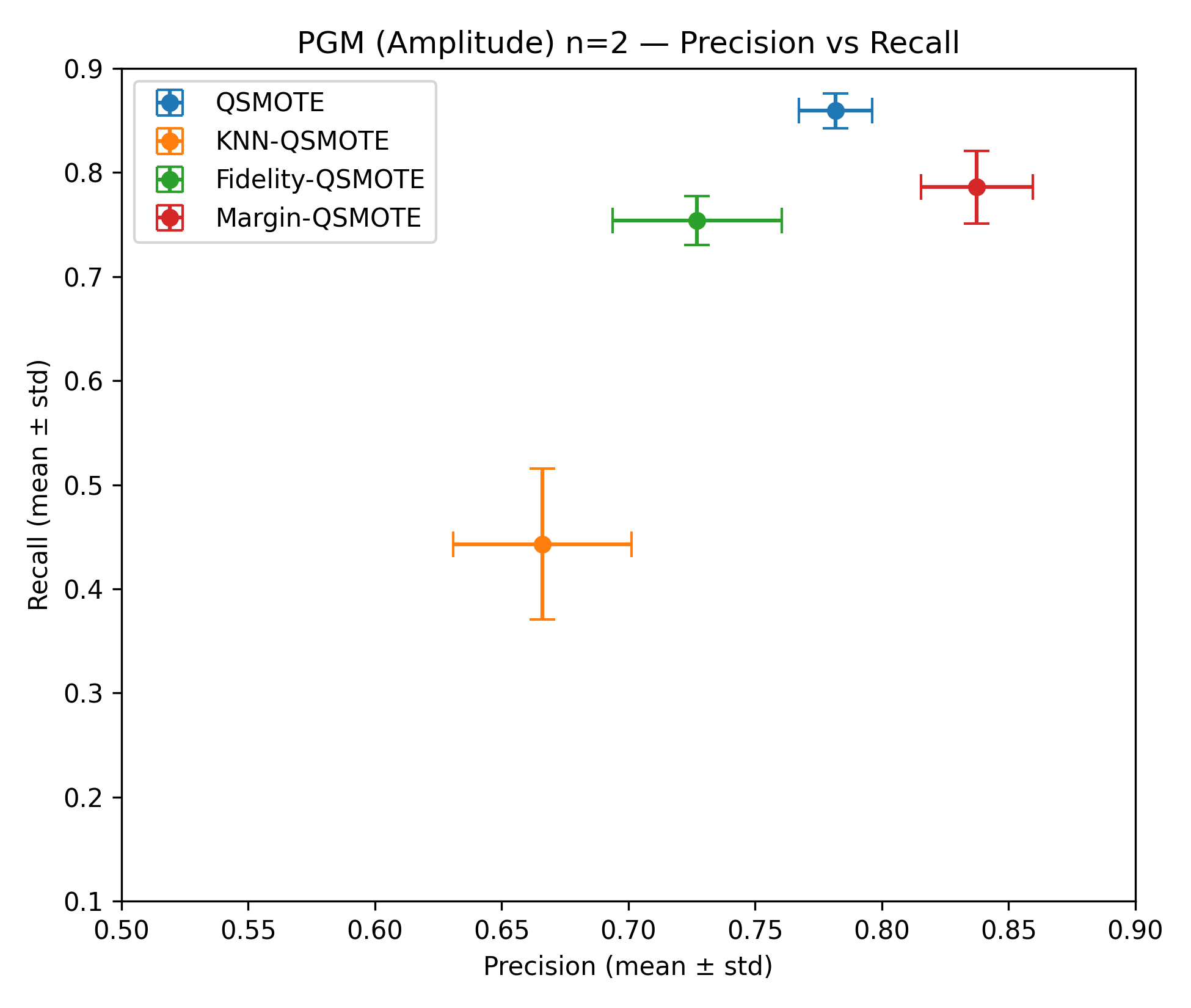}
    \caption{PGM (Amplitude), $n\_copies=2$}
    \label{fig:pgm_amp_n2_pr}
  \end{subfigure}

  \vspace{0.8em}

  \begin{subfigure}{0.48\textwidth}
    \centering
    \includegraphics[width=0.8\linewidth]{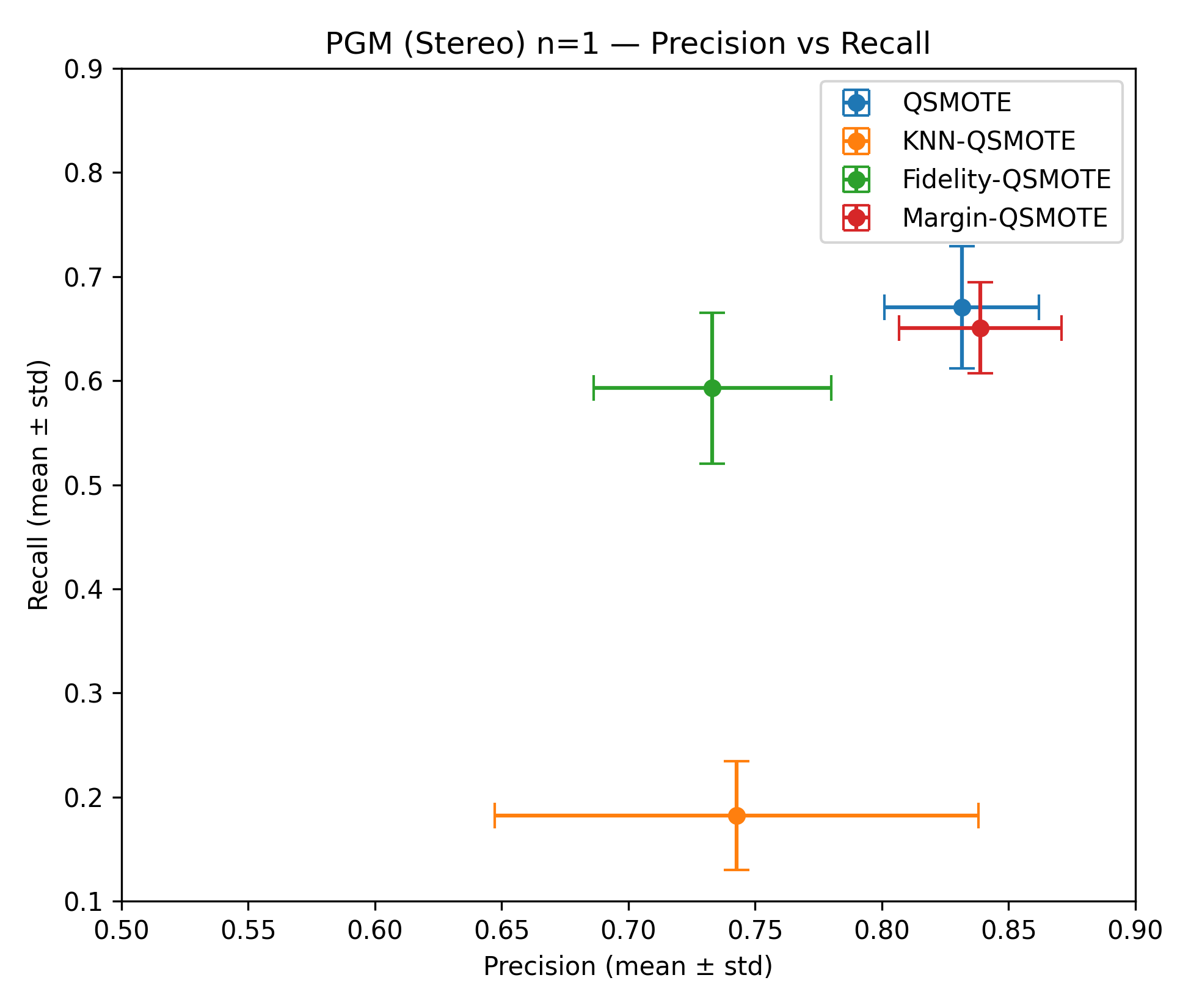}
    \caption{PGM (Stereo), $n\_copies=1$}
    \label{fig:pgm_st_n1_pr}
  \end{subfigure}\hfill
  \begin{subfigure}{0.48\textwidth}
    \centering
    \includegraphics[width=0.8\linewidth]{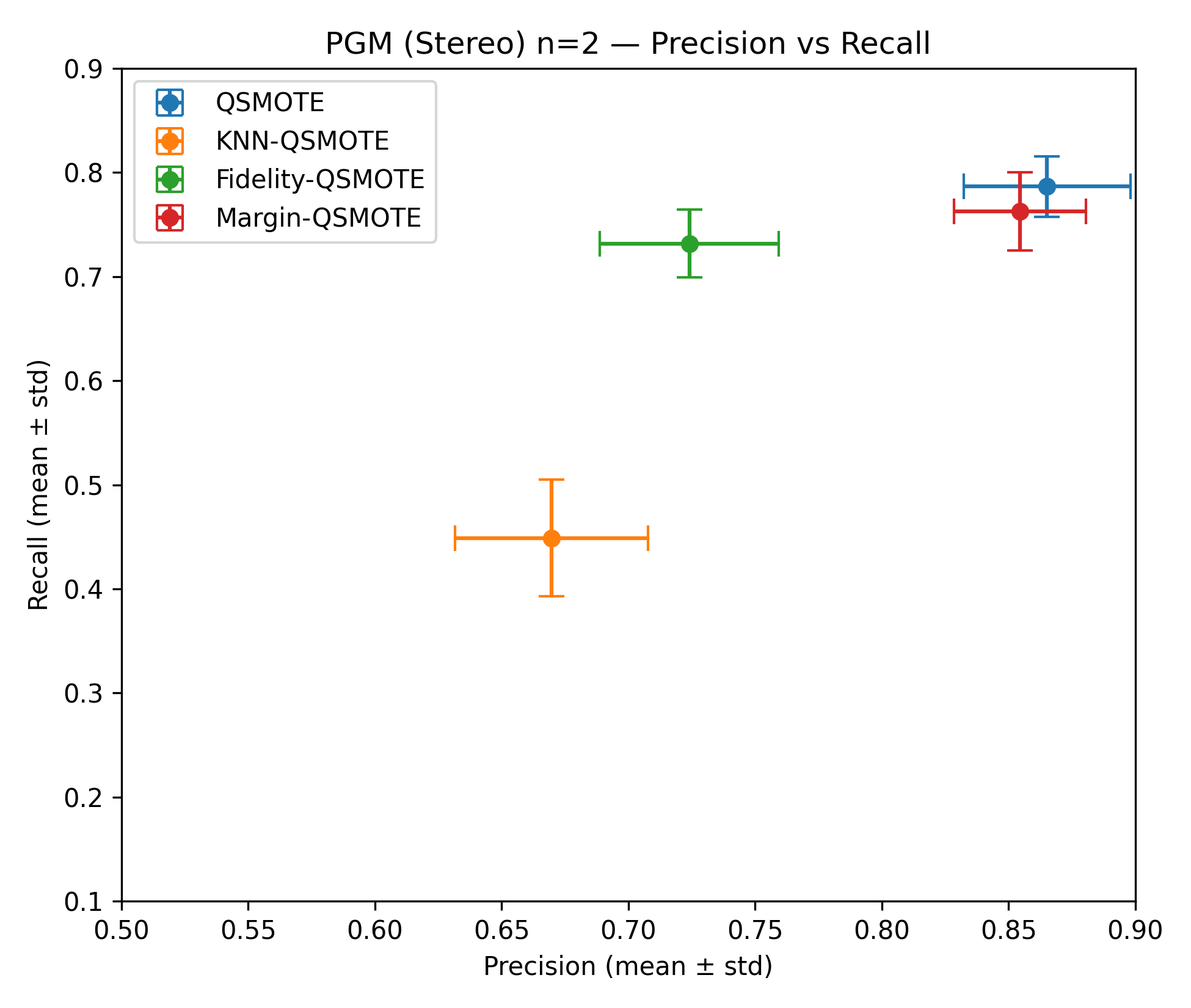}
    \caption{PGM (Stereo), $n\_copies=2$}
    \label{fig:pgm_st_n2_pr}
  \end{subfigure}
  \caption{Precision-Recall tradeoffs for PGM under amplitude and stereo encodings with $n\_copies\in\{1,2\}$.}
  \label{fig:pgm_pr_grid}
\end{figure*}

\begin{figure*}[htbp]
  \centering
  \begin{subfigure}{0.48\textwidth}
    \centering
    \includegraphics[width=0.8\linewidth]{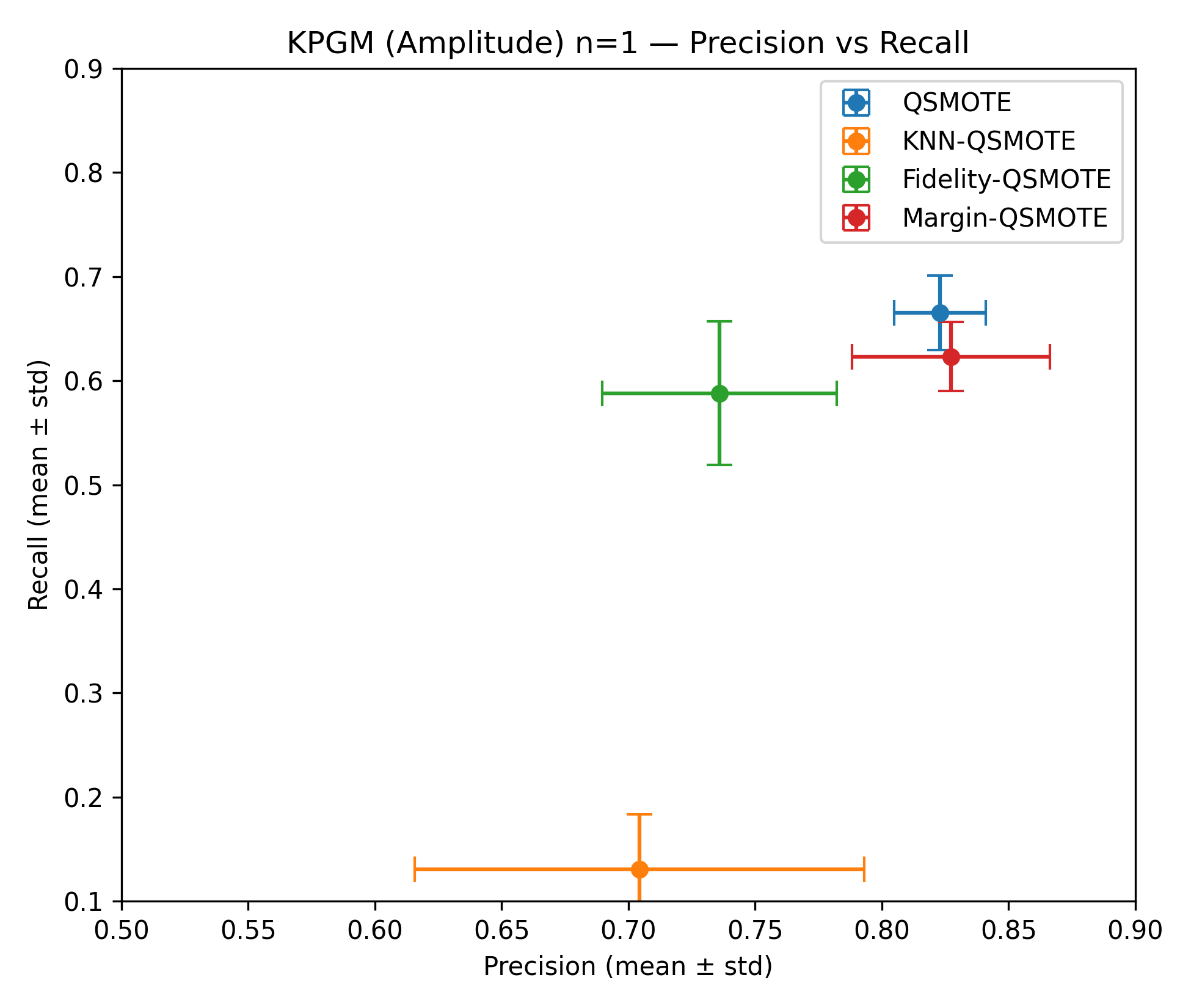}
    \caption{KPGM (Amplitude), $n\_copies=1$}
    \label{fig:kpgm_amp_n1_pr}
  \end{subfigure}\hfill
  \begin{subfigure}{0.48\textwidth}
    \centering
    \includegraphics[width=0.8\linewidth]{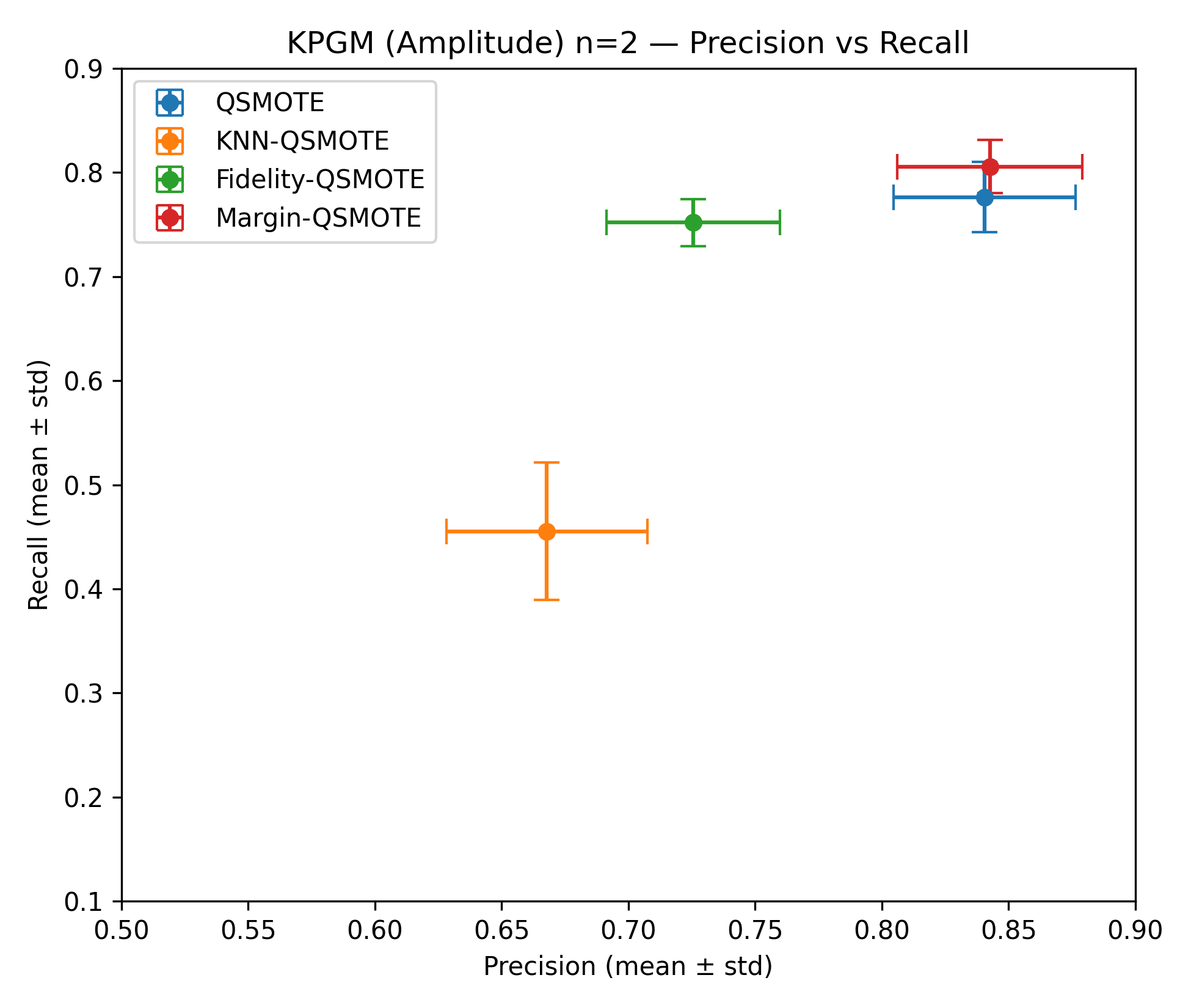}
    \caption{KPGM (Amplitude), $n\_copies=2$}
    \label{fig:kpgm_amp_n2_pr}
  \end{subfigure}

  \vspace{0.8em}

  \begin{subfigure}{0.48\textwidth}
    \centering
    \includegraphics[width=0.8\linewidth]{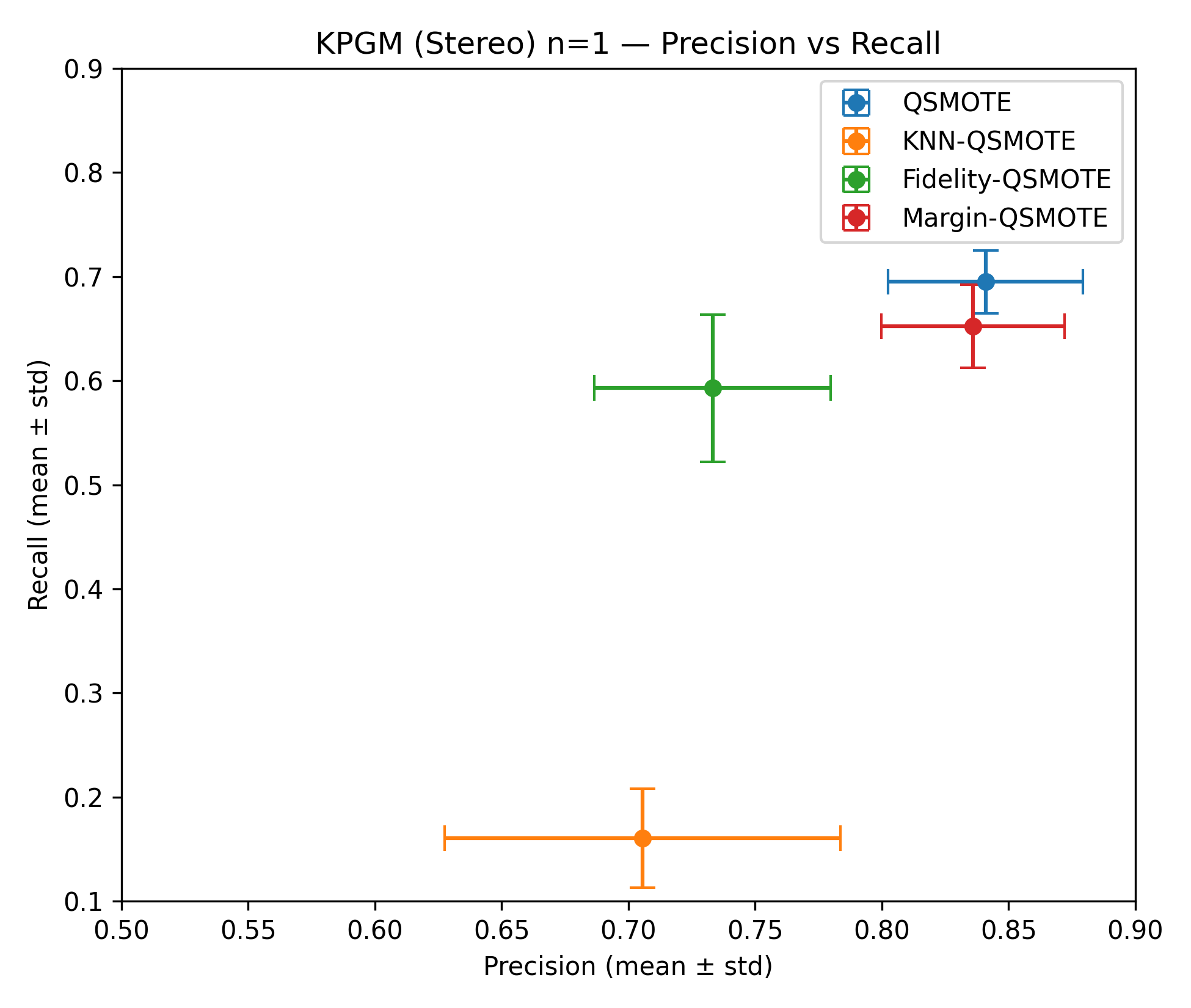}
    \caption{KPGM (Stereo), $n\_copies=1$}
    \label{fig:kpgm_st_n1_pr}
  \end{subfigure}\hfill
  \begin{subfigure}{0.48\textwidth}
    \centering
    \includegraphics[width=0.8\linewidth]{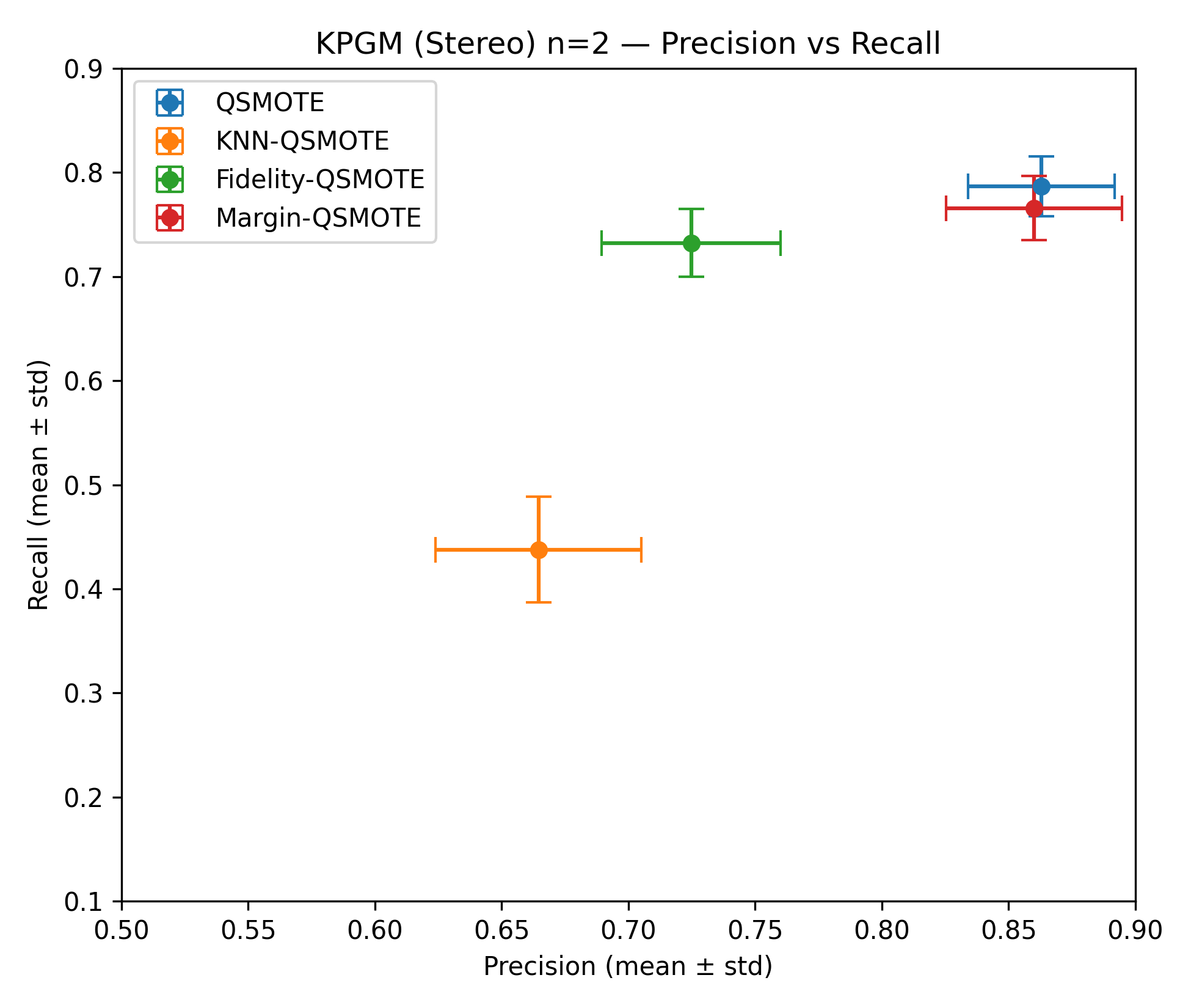}
    \caption{KPGM (Stereo), $n\_copies=2$}
    \label{fig:kpgm_st_n2_pr}
  \end{subfigure}
  \caption{Precision-Recall tradeoffs for KPGM under amplitude and stereo encodings with $n\_copies\in\{1,2\}$.}
  \label{fig:kpgm_pr_grid}
\end{figure*}

Table~\ref{tab:qsmote_kpgm_stereo} presents the results of QSMOTE and its variants with the KPGM classifier using stereo encoding. For $n\_copies=1$, QSMOTE outperformed all other variants with an accuracy of $0.8078 \pm 0.0176$ and an F1-score of $0.7602 \pm 0.0218$, reflecting its ability to achieve both high precision and recall. Margin-QSMOTE also performed competitively, with slightly lower accuracy but strong precision, while Fidelity-QSMOTE showed moderate balanced performance. In contrast, KNN-QSMOTE again underperformed severely, with recall as low as $0.1603 \pm 0.0475$, resulting in the weakest F1-score. When $n\_copies=2$ were used, the results improved across most variants, with QSMOTE achieving the highest accuracy ($0.8511 \pm 0.0165$) and F1-score ($0.8225 \pm 0.0189$). Margin-QSMOTE followed closely, with balanced precision and recall and an F1-score of $0.8094 \pm 0.0204$. Fidelity-QSMOTE remained consistent but less competitive, and KNN-QSMOTE improved modestly but continued to lag. Overall, stereo encoding with additional copies strengthened classifier performance, with QSMOTE and Margin-QSMOTE emerging as the most effective strategies under the KPGM framework. The comparative analysis across classifiers highlights clear advantages of the quantum approaches over the classical baseline. With RF, the best performance was observed for Margin-QSMOTE, achieving an accuracy of $0.8371$ and F1-score of $0.7516$, while Fidelity-QSMOTE was close in terms of F1-score ($0.7785$). However, the classical model consistently exhibited lower recall compared to the quantum classifiers, limiting its ability to capture minority class instances effectively. In contrast, the PGM classifier demonstrated stronger performance, particularly when $n\_copies=2$ were employed. Under amplitude encoding, Margin-QSMOTE yielded an F1-score of $0.8102$, while in stereo encoding, the QSMOTE baseline outperformed other variants with accuracy of $0.8512$ and F1-score of $0.8234$. A notable advantage of PGM was its ability to achieve very high recall, reaching $0.8594$ for the amplitude baseline. Similarly, the KPGM classifier produced competitive results, often surpassing PGM in precision. Its best outcomes were recorded with $n\_copies=2$, where stereo encoding achieved accuracy of $0.8511$ and F1-score of $0.8225$, while amplitude encoding reached accuracy of $0.8483$ and F1-score of $0.8230$. Overall, both quantum classifiers consistently outperform RF, with stereo-encoded QSMOTE under PGM and KPGM emerging as the most effective configurations. The overall comparison between RF, PGM, and KPGM classifiers underscores the advantages of quantum-inspired approaches for imbalanced learning with QSMOTE. RF provides a decent classical baseline but was limited by its relatively low F1-scores ($\leq$ 0.78) and reduced recall, indicating weaker handling of the minority class. In contrast, the PGM classifier demonstrated a clear improvement, with amplitude encoding favoring margin-based QSMOTE and stereo encoding favoring the baseline QSMOTE variant. The use of $n\_copies=2$ consistently amplified recall, making PGM particularly effective in capturing minority instances. The KPGM classifier performed comparably to PGM while exhibiting greater stability across QSMOTE variants. Its top results were achieved with $n\_copies=2$, yielding accuracy and F1-scores of $0.8511$ and $0.8225$ (stereo encoding) and $0.8483$ and $0.8230$ (amplitude encoding), respectively. Overall, both PGM and KPGM with $n\_copies=2$ substantially outperformed the RF baseline, with stereo-encoded QSMOTE emerging as the most effective configuration in terms of accuracy and balanced F1 performance.


The RF baseline in Fig.~\ref{fig:rf_overview} shows that among QSMOTE variants, Fidelity-QSMOTE achieves the highest F1 ($0.7785\pm0.0298$), followed by Margin-QSMOTE ($0.7516\pm0.0327$) and QSMOTE ($0.7457\pm0.0268$), while KNN-QSMOTE lags behind at $0.5743\pm0.0324$. The corresponding precision-recall scatter further clarifies this result: KNN-QSMOTE suffers from extremely low recall ($0.5224\pm0.0584$), whereas Fidelity-QSMOTE balances both precision ($0.7647\pm0.0412$) and recall ($0.7940\pm0.0300$), explaining its superior F1 performance. For the PGM classifier under amplitude encoding, Fig.~\ref{fig:pgm_amp_f1_bars} shows that increasing the number of copies from $n\_copies=1$ to $n\_copies=2$ consistently boosts F1 across all variants. QSMOTE improves from $0.7166\pm0.0366$ to $0.8185\pm0.0069$, Fidelity-QSMOTE from $0.6503\pm0.0382$ to $0.7395\pm0.0155$, Margin-QSMOTE from $0.7398\pm0.0287$ to $0.8102\pm0.0197$, and KNN-QSMOTE from $0.1917\pm0.0633$ to $0.5273\pm0.0487$. The improvements stem largely from recall gains, as seen in QSMOTE recall rising from $0.6384\pm0.0595$ to $0.8594\pm0.0168$. A similar trend is observed with stereo encoding in Fig.~\ref{fig:pgm_stereo_f1_bars}, where QSMOTE increases from $0.7407\pm0.0361$ to $0.8234\pm0.0235$, Margin-QSMOTE from $0.7313\pm0.0235$ to $0.8053\pm0.0231$, and Fidelity-QSMOTE from $0.6516\pm0.0402$ to $0.7269\pm0.0183$. KNN-QSMOTE also benefits, rising from $0.2876\pm0.0693$ to $0.5343\pm0.0381$. Comparing encodings at $n\_copies=2$, stereo yields a slightly higher maximum F1 ($0.8234$ vs.~$0.8185$), suggesting a modest advantage. The line plots in Fig.~\ref{fig:pgm_effect_copies} highlight the consistent benefit of using $n\_copies=2$ for PGM across both amplitude and stereo encodings. The strongest relative improvements occur for KNN-QSMOTE, where recall deficiencies at $n\_copies=1$ are alleviated by additional copies, confirming the role of repeated preparations in stabilizing quantum measurements. For KPGM with amplitude encoding, Fig.~\ref{fig:kpgm_amp_f1_bars} shows that Margin-QSMOTE achieves the best F1 at $n\_copies=2$ ($0.8230\pm0.0197$), followed closely by QSMOTE ($0.8061\pm0.0205$). Fidelity-QSMOTE remains moderate ($0.7377\pm0.0163$), while KNN-QSMOTE, although improved, stays relatively weak ($0.5373\pm0.0439$). With stereo encoding (Fig.~\ref{fig:kpgm_stereo_f1_bars}), the leaders are again QSMOTE ($0.8225\pm0.0189$) and Margin-QSMOTE ($0.8094\pm0.0204$), which outperform Fidelity- and KNN-QSMOTE. The comparison between encodings indicates that at $n\_copies=2$, both amplitude and stereo achieve similar top F1 values, showing robustness across encodings. The effect of copy number for KPGM is further illustrated in Fig.~\ref{fig:kpgm_effect_copies}, where all variants gain from $n\_copies=2$. The largest improvements again occur for KNN-QSMOTE, with F1 rising from $0.2153$ to $0.5373$ under amplitude and from $0.2565$ to $0.5248$ under stereo. This confirms that multiple copies mitigate recall weakness in this variant. The precision-recall grids in Figs.~\ref{fig:pgm_pr_grid} and~\ref{fig:kpgm_pr_grid} further substantiate these findings. For PGM, QSMOTE and Margin-QSMOTE clearly occupy the upper-right region at $n\_copies=2$, combining high precision (up to $0.8652\pm0.0328$) with improved recall (up to $0.7865\pm0.0291$). KNN-QSMOTE shifts upward compared to $n\_copies=1$, but still trails. Similarly, KPGM precision-recall plots show QSMOTE and Margin-QSMOTE moving towards balanced precision-recall performance, while KNN-QSMOTE remains biased towards precision with weaker recall. Overall, across classifiers and encodings, $n\_copies=2$ systematically lifts recall while preserving precision, producing consistent F1 improvements. Taken together, the results demonstrate three key points: first, increasing the number of copies substantially improves quantum classifiers, primarily by enhancing recall; second, QSMOTE and Margin-QSMOTE are the most reliable oversampling strategies in the quantum setting, often surpassing the RF baseline; and third, amplitude and stereo encodings yield comparable performance at $n\_copies=2$, suggesting that copy number is the more influential factor compared to encoding choice.

\subsection{Discussion}
The comparative evaluation across RF, PGM, and KPGM classifiers provides several key insights into the role of quantum-inspired methods in handling imbalanced learning scenarios. First, the RF baseline confirmed its strength as a widely used classical model but revealed inherent limitations in recall and minority class recognition. Despite producing reasonable accuracy and precision, its F1-scores ($\leq 0.78$) underscored the challenge of balancing performance in skewed datasets. In contrast, the quantum-inspired classifiers consistently outperformed RF, demonstrating the utility of leveraging Hilbert space geometry for classification. The PGM classifier, in particular, showed substantial improvements when multiple quantum copies were employed. Under amplitude encoding, margin-based QSMOTE yielded strong results, while stereo encoding favored the baseline QSMOTE variant. With $n\_copies=2$, PGM achieved remarkable recall ($0.8594$ in amplitude) and overall F1-scores exceeding $0.82$, highlighting its capacity to capture minority class patterns that classical ensembles missed. The KPGM classifier delivered competitive results and, in some cases, outperformed PGM in precision. Its most notable strength was stability across QSMOTE variants, where amplitude and stereo encoding with $n\_copies=2$ both yielded accuracies above $0.84$ and F1-scores above $0.82$. This robustness suggests that kernelized formulations inherit the scalability and generalization properties of kernel methods while retaining the benefits of quantum-inspired similarity measures. Across both classifiers, two consistent trends emerged. First, increasing the number of copies from one to two systematically improved recall and F1 performance, confirming the theoretical advantage of leveraging additional quantum samples. Second, encoding choice played a decisive role: amplitude encoding enhanced recall and margin-based sampling, whereas stereo encoding favored the baseline QSMOTE approach, delivering the highest overall accuracy and F1 performance. Together, these findings demonstrate that quantum-inspired classifiers not only provide measurable gains over classical baselines but also offer complementary strengths: PGM excels in recall and encoding-dependent optimizations, while KPGM offers stability and balance across sampling strategies. This duality indicates that the choice between PGM and KPGM should be guided by dataset characteristics, encoding feasibility, and the desired trade-off between precision and recall.

\section{Conclusion \label{SecV}}
In this work, we presented a unified investigation of quantum-inspired classifiers grounded in kernel methods and quantum measurement theory, with particular focus on the interplay between QSMOTE-based data augmentation and classifier performance. By benchmarking against a RF baseline, we demonstrated that quantum-inspired approaches, namely PGM and KPGM, consistently deliver superior classification outcomes, especially in terms of recall and balanced F1-scores. The empirical analysis revealed that while RF offered a reasonable baseline, it struggled with minority class detection, highlighting the limitations of classical ensembles for imbalanced datasets. In contrast, PGM classifiers benefited significantly from multiple quantum copies, with stereo encoding and $n\_copies=2$ achieving the highest performance overall. KPGM classifiers, though slightly less sensitive to encoding choices, proved more stable across QSMOTE variants, yielding competitive top scores under both amplitude and stereo encodings. These results underline two key insights: first, quantum-inspired classifiers can be tuned through encoding strategies and copy number to achieve notable performance gains; second, KPGM’s robustness makes it a promising candidate for broader practical deployment where dataset characteristics vary. Taken together, our theoretical and empirical findings provide a comprehensive perspective on the complementarity of kernel-based and measurement-based QiML approaches, offering clear guidance for their application in imbalanced learning tasks and beyond.

\backmatter
\section*{Declarations}

\begin{itemize}
\item Conflict of interest/Competing interests: The authors have no financial or non-financial competing interests.
\item Author contribution: The authors confirm their contribution to the paper as
follows: Study conception, design and methodology: B.K.B., G.S., R.G.;
Data collection: B.K.B.;
Analysis and interpretation of results: B.K.B., G.S., R.G.;
Draft manuscript preparation: B.K.B., G.S., R.G.;
Supervision: G.S., R.G.;
All authors reviewed the results and approved the final version of the manuscript.
\item Data Availability: data, algorithms, and all materials necessary to reproduce and verify the experimental results reported in this manuscript are available from the corresponding author upon reasonable request.

\end{itemize}

\noindent

\bibliography{sn-article}

\end{document}